\documentclass[lettersize,journal]{IEEEtran}
\usepackage{amsmath,amsfonts}
\usepackage{algorithmic}
\usepackage{algorithm}
\usepackage{array}
\usepackage[caption=false,font=normalsize,labelfont=sf,textfont=sf]{subfig}
\usepackage{textcomp}
\usepackage{stfloats}
\usepackage{url}
\usepackage{verbatim}
\usepackage{graphicx}
\usepackage{cite}
\usepackage{enumerate}
\usepackage{booktabs}
\usepackage{multirow}
\usepackage{bm}
\usepackage{tikz}
\usetikzlibrary{decorations.markings}
\usetikzlibrary{matrix, arrows.meta, bending}

\newtheorem{theorem}{Theorem}

\newtheorem{definition}{Definition}
\newtheorem{example}{Example}

\begin{document}

\title{
Continuous-tone Simple Points: An $\ell_0$-Norm of Cyclic Gradient for Topology-Preserving Data-Driven Image Segmentation
}

\author{Wenxiao Li, Faqiang Wang, Yuping Duan, Li Cui, Liqiang Zhang and Jun Liu
\thanks{This work was supported by the National Natural Science Foundation of China under Grant 42293272 and 12371527. (Corresponding author: Jun Liu.)

Wenxiao Li, Faqiang Wang, Yuping Duan, Li Cui, Jun Liu are with the Laboratory of Mathematics and Complex Systems (Ministry of Education), School of Mathematical Sciences, Beijing Normal University, Beijing, 100875, People's Republic of China (e-mails: wxli@mail.bnu.edu.cn, fqwang@bnu.edu.cn, licui@bnu.edu.cn, ypduan@bnu.edu.cn, jliu@bnu.edu.cn).

Liqiang Zhang is with the State Key Laboratory of Remote Sensing Science, Faculty of Geographical Science, Beijing Normal University, Beijing, 100875, People's Republic of China (e-mail: zhanglq@bnu.edu.cn).}
}

\markboth{Journal of \LaTeX\ Class Files,~Vol.~14, No.~8, August~2021}%
{Shell \MakeLowercase{\textit{et al.}}: A Sample Article Using IEEEtran.cls for IEEE Journals}


\maketitle

\begin{abstract}
Topological features play an essential role in ensuring geometric plausibility and structural consistency in image analysis tasks such as segmentation and skeletonization. However, integrating topology-preserving learning based on simple points into deep learning tasks remains challenging, as existing simple point detection methods are confined to binary images and are non-differentiable, rendering them incompatible with gradient-based optimization in modern deep learning. Moreover, morphological and purely data-driven approaches often fail to guaranty topological consistency. To address these limitations, we propose a novel method that directly computes simple points on continuous-valued images, enabling differentiable topological inference. Building on this theory, we develop an efficient skeleton extraction algorithm that preserves topological structures in binary and continuous-valued images. Furthermore, we design a variational model that enforces topological constraints by preserving topologically non-removable (i.e., non-simple) points, which can be seamlessly integrated into any deep neural network segmentation with softmax or sigmoid outputs. Experimental results demonstrate that the proposed approach effectively improves topological integrity and structural accuracy across multiple benchmarks.
\end{abstract}


\section{Introduction}
\IEEEPARstart{T}{opological} information captures key structural features and spatial relationships within images, playing an indispensable role in various visual fields such as medical imaging and remote sensing. In medical image analysis, the topological properties (e.g. connectivity, genus) of tissue structures, vascular networks, and lesion regions are critical for disease diagnosis, quantification, and progression tracking. In remote sensing, the topological characteristics of land cover (e.g., connectivity of road networks, integrity of land parcel shapes) directly affect the precision of geographical information extraction and scene understanding. Integrating topological constraints into tasks such as image segmentation can significantly enhance the geometric plausibility and structural consistency of results, avoiding fragmented or topologically erroneous regions. 

\begin{figure}[!t]
\centering
\includegraphics[width=3.45in]{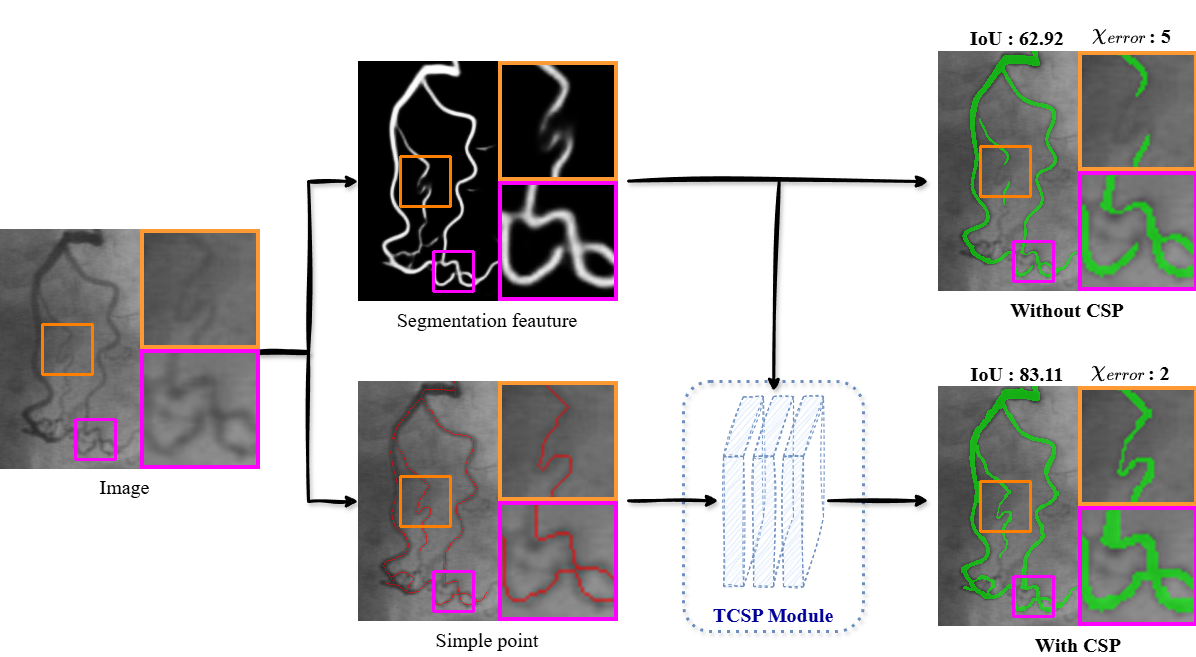}
\caption{Comparison of segmentation results from SAM2 \cite{sam2} without and with the proposed continuous-tone simple points (CSP) method. The image in the right column is a partial enlarged view of the area with the same color border in the left column image.}
\label{fig_main}
\end{figure}

With the growth of data volumes and advances in hardware, data-driven image segmentation methods have developed rapidly. Although these models \cite{deeplabv3+,unet,unet++,segformer,sam, sam2} have achieved success in attaining high pixel-level accuracy, they typically treat segmentation as a pixel-wise classification problem. This perspective struggles to enforce global topological properties of segmented objects, which can adversely affect downstream tasks in fields such as medicine and remote sensing. Consequently, numerous scholars have made outstanding contributions to addressing these challenges. The first approaches improved the model's ability to perceive fine details through methods such as snake convolution \cite{snakeconv}, fractal geometry \cite{fractal}, Euler characteristic \cite{euler}, discrete Morse theory \cite{dmtloss, dmttopologyaware}, the extraction of intermediate-layer features from VGG \cite{vggtopo}, and learning based method \cite{structure-agnostic}, etc.. These techniques improved the representation of particularly challenging structures, thus boosting the segmentation performance of tubular objects. Subsequently, nonlinear relationships between adjacent pixels were established to improve the connectivity of the segmentation results. Representative works in this direction included the morphological skeleton \cite{cldice,learnmorph,clce}, the construction of connectivity matrices \cite{biconnet,roadnet,conn}, the max-min affinity learning \cite{affinity}, among others. By imposing soft constraints, these approaches improved segmentation connectivity and encouraged the network to focus on structural characteristics. There were also methods that integrated the concept of image registration with segmentation, enforcing the topological consistency between the segmentation results and the manual initial values by the Beltrami coefficient \cite{beltrami}, the Jacobian matrix \cite{ZHANG2021218, hyperelastic, quasiconformal}, or the approaches based on DARTEL (Diffeomorphic Anatomical Registration using Exponentiated Lie algebra) \cite{deformation, diffeomorphic}. However, due to their sensitivity to initial values, these methods were not well-suited for complex tubular and reticular structures, such as blood vessels. Another category of methods used persistent homology \cite{hutopoloss,cloughtopoloss,postph,widthtopo,topocellgen} from topological data analysis \cite{gudhi}, which improved the results by imposing the consistency of Betti numbers with segmentation and ground-truth. Nevertheless, both types of approach entailed substantial computational overhead, rendering them impractical in data-driven modeling frameworks. Similarly in spirit to registration-based approaches, methods based on the concept of simple points \cite{topolevelset,toposegnet,topostd,deepclose,spskel} preserved topology by selectively removing or adding points in a digital image that did not alter its global topological structure, thereby ensuring that topological consistency was preserved during the iterative process. 

However, deep learning methods produce probabilistic outputs and are optimized via gradient-based techniques, whereas existing simple point computation methods are restricted to binary images and are therefore not directly applicable. Thus, we propose a method capable of directly computing simple points on continuous-valued images. This approach enables the skeleton to be extracted from continuous-valued images and can be seamlessly integrated into data-driven modeling frameworks, as shown in Fig.\ref{fig_main}, allowing gradient backpropagation to update the model. The main contributions of this paper are summarized as follows:
\begin{itemize}
    \item We propose a method for computing simple points on continuous-valued image features. This method effectively extracts simple point information from the predicted probability maps of semantic segmentation networks, i.e., the topological features of the image. The smoothness of this method theoretically ensures that backpropagation can be performed stably after introducing the features of continuous-tone simple points into the architectural design of image segmentation networks.
    \item Based on the theory of continuous-tone simple points (CSP), we design a skeleton extraction algorithm that achieves higher computational efficiency while preserving topological consistency on both binary and continuous-valued images and that can be easily integrated into deep learning pipelines.
    \item We propose a variational model based on continuous simple point constraints, which enforces the segmentation result to preserve topologically non-removable points (i.e., non-simple points), thereby ensuring topological consistency. Moreover, this model can be integrated into any segmentation network equipped with softmax or sigmoid activation functions (TCSP module of Fig. \ref{fig_all}).
\end{itemize}

The remainder of this paper is structured as follows. Section \ref{sec_relate} surveys previous work closely related to the proposed approaches. Section \ref{sec_method} details the proposed method for computing continuous-ton simple points and describes its application. Section \ref{sec_experiment} provides the comprehensive performance evaluation to investigate effectiveness. Finally, Section \ref{sec_conclusion} concludes this paper.

\section{Related Work}\label{sec_relate}
\subsection{Topologically Simple Points in Binary Images}

A digital image is typically modeled as a function defined on a discrete grid (usually in two or three dimensions). Let $\mathbb{Z}^d$ be the $d$-dimensional integer space (for 2D images, $d=2$) and the domain of the image $\Omega$ is a finite subset of $\mathbb{Z}^d$. Typically, for a 2D image, the  coordinate of a pixel is $\bm x = (x_1,x_2) \in \Omega \subset \mathbb{Z}^2$. Due to the grid nature of the image, we define $\mathbb{N}_n(\bm x)$ as the $n$-neighborhood of a point $\bm x \in \Omega$, where $n \in \{ 4,8\}$ for the 2D case:
\begin{align*}
    \mathbb{N}_4(\bm x) &= \{ \bm x' = (x_1',x_2'): (|x_1-x_1'| + |x_2-x_2'|) \le 1) \}, \\
    \mathbb{N}_8(\bm x) &= \{ \bm x' = (x_1',x_2'): \max(|x_1-x_1'|, |x_2-x_2'|) \le 1) \}. 
\end{align*}
 Similarly, for the 3D case, $\mathbb{N}_n(\bm x)$ contains the 6, 18, or 26 neighbors of $\bm x$ and $\bm x$ itself, depending on the value of $n$. In addition, we denote $\mathring{\mathbb{N}}_n(\bm x) = \mathbb{N}_n(\bm x)\setminus \bm x$. In the following, unless otherwise specified, we use the non-bold $x$ to denote the coordinate of the domain $\Omega$.

Next, define the pixel value at each coordinate point as $u(x)$. If the range of $u(x)$ is $\{0,1\}$, the image is termed a binary image. The set $\mathbb{X} = \{ x \in \Omega: u(x)=1 \}$ is called the foreground and its complement $ \mathbb{X}^{c} = \Omega \setminus \mathbb{X}$ is called the background. To avoid topological paradoxes, complementary adjacency relations must be specified for the foreground and background. We emphasize that throughout this paper we consider the scenario where the foreground is 8-connected and the background is 4-connected. The following formal definitions are from \cite{sp_topo}.
\begin{definition}[Geodesic Neighborhood \cite{sp_topo}] \label{def_neighbor}
Let $\mathbb{X} \subset \Omega$ and $x \in \Omega$. The geodesic neighborhood of $x$ with respect to $\mathbb{X}$ of order $k$ is the set $\mathbb{N}^k_n(x, \mathbb{X})$ defined recursively by:
\begin{align*}
    \mathbb{N}^1_n(x, \mathbb{X}) &= \mathring{\mathbb{N}}_n(x) \cap \mathbb{X}, \\
    \mathbb{N}^k_n(x, \mathbb{X}) &= \cup\{\mathbb{N}_n(y) \cap \mathring{\mathbb{N}}_m(x) \cap \mathbb{X}, y \in \mathbb{N}^{k-1}_n(x, \mathbb{X})\},
\end{align*}
where $n=4$, $m=8$ in 2D.
\end{definition}
\begin{definition}[Topological Numbers \cite{sp_topo}]
Let $\mathbb{X} \subset \Omega$ and $x \in \Omega$. The topological numbers of the point $x$ relative to the set $\mathbb{X}$ are: 
\begin{align*}
    T_4(x, \mathbb{X}) &= \# \mathbb{T}_4(\mathbb{N}^{2}_4(x,\mathbb{X})), \\
    T_8(x, \mathbb{X}) &= \# \mathbb{T}_8(\mathbb{N}^{1}_8(x,\mathbb{X})),
\end{align*}
where $\mathbb{T}_n(\mathbb{X})$ represents the set of all n-connected components of $\mathbb{X}$ and $\#$ denotes the cardinality of a set. 
\end{definition}

Intuitively, an n-connected neighbor of point $x$ lies in its geodesic neighborhood $\mathbb{N}^{k}_{n}(x, \mathbb{X})$ if there exists a path in $\mathbb{X}$ of maximum length $k$ between $x$ and its neighbors \cite{topolevelset}. In addition, topological numbers are used to classify the topology type of a grid point, especially for the characterization of simple points. A point $x \in \Omega$ is referred to as a simple point if and only if changing its pixel value $u(x)$ (from 0 to 1 or from 1 to 0) does not alter the global topology of the image.
\begin{definition}[Simple point \cite{sp_topo}] \label{def_sp}
For $\mathbb{X} \subset \Omega$ and $\mathbb{X}^{c} = \Omega \setminus \mathbb{X}$, a point $x\in \Omega$ is called a simple point if and only if $T_4(x,\mathbb{X}^{c})=1$ and $T_8(x, \mathbb{X})=1$. 
\end{definition}

\begin{figure}[!t]
\centering
\includegraphics[width=3.2in]{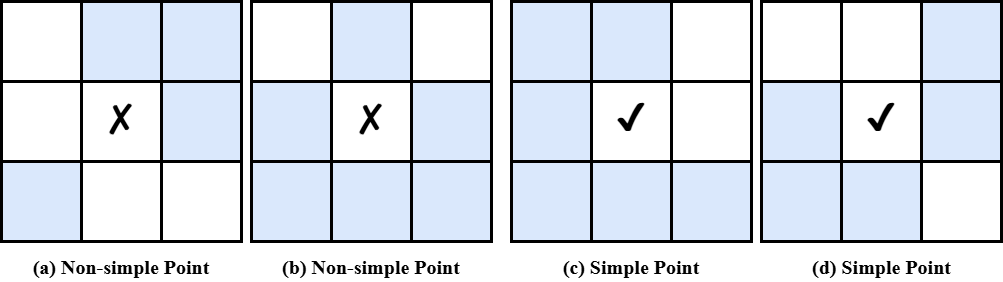}
\caption{Examples of non-simple and simple points. The solid grids means foreground and the background is hollow.
Changing the value of point $x$ alters the topological properties in cases (a) and (b), but does not affect the topology in cases (c) and (d).}
\label{fig_exsp}
\end{figure}

For a more intuitive understanding, we provide the following example.
\begin{example}[The illustration of the simple point]
    In Fig. \ref{fig_exsp}, $\Omega$ is the set of all grids, the foreground $\mathbb{X}$ is the set of all solid grids and the background $\mathbb{X}^{c}$ is hollow; we can calculate the topology number of the center point $x$ in each case:
    \begin{enumerate}[(a)]
        \item $T_8(x,\mathbb{X})=2$ and $T_4(x, \mathbb{X}^{c})=2$: non-simple point.
        \item $T_8(x,\mathbb{X})=1$ and $T_4(x, \mathbb{X}^{c})=0$: non-simple point.
        \item $T_8(x,\mathbb{X})=1$ and $T_4(x, \mathbb{X}^{c})=1$: simple point.
        \item $T_8(x,\mathbb{X})=1$ and $T_4(x, \mathbb{X}^{c})=1$: simple point.
    \end{enumerate}
\end{example}

Due to its property of preserving topological invariance during iterative processes, the theory of simple points has been widely applied in fields such as image segmentation \cite{topolevelset,topostd}, image inpainting \cite{deepclose}, and other fields. However, the fact that it can only determine simple points in binary images imposes significant limitations when applied to the probabilistic feature maps, the output of deep learning models.

\subsection{Skeletonization Algorithms and Applications}

Skeletonization algorithms aim to extract the medial axis of an object, which encapsulates the geometric and topological features of the image. Consequently, skeleton features have been widely applied in computer vision tasks such as image description, reconstruction, recognition, generation, and segmentation \cite{MA1996420,MORSE1994327,deepclose,cldice,ZHAO20071270,Thibault2000}. To achieve better performance in practice, \cite{Bertrand2014,zhangskel,795212,SAHA19971939,1114851,5396343} explored and discussed numerous notable efforts to improve the accuracy of skeleton features in discrete digital image spaces. Furthermore, \cite{skel_survey} provided a comprehensive overview and classification of skeletonization algorithms and their applications, facilitating subsequent researchers in their study and understanding.

In the current era, the use of data-driven deep learning methods to address computer vision tasks has become mainstream. However, traditional skeletonization methods are incompatible with such approaches, which rely on gradient backpropagation for parameter optimization. Consequently, to integrate skeletal features into deep learning networks, many researchers have made interesting attempts. The first category of work is based on mathematical morphology \cite{cldice}. By employing the max-pooling and the min-pooling kernels to simulate dilation and erosion operators, morphological skeletons were extracted. Nevertheless, morphological skeletons are often fragmented, causing the extracted skeleton to deviate from the true medial axis in terms of geometry and topology. Subsequently, several studies have improved upon the disconnection issue of morphological skeletons. The most mathematically principled approach being the smoothing of erosion and dilation operators \cite{learnmorph}. The second category encompassed fully data-driven methods \cite{8000414, 8099896, 9178497}. Most of these works trained encoder-decoder networks using paired input images and ground-truth skeletons obtained via classical skeletonization algorithms. However, these learning-based methods cannot guaranty topological consistency and are susceptible to domain shifts between training and inference data.

To address the aforementioned problem where gradient-optimizable skeletons may compromise the topological properties of objects, \cite{spskel} proposed an inspiring idea that bridges the gap between traditional skeletonization principles (with strict topological guaranties) and gradient-based optimization methods. By iterative removal of topologically simple points, the extracted skeleton is constrained to maintain unchanged topological properties. However, topologically simple points are defined in binary images, which conflict with the probability maps output by networks. To resolve this, the paper employed a reparameterization trick \cite{reparam} and utilized the straight-Through Estimator (STE) \cite{ste} to enable gradient-based optimization. Nevertheless, this approach still had several problems. First, introducing steps such as reparameterization and repeated sampling will increase computational overhead and be sensitive to hyperparameters, resulting in limited generalization capability. Second, random noise introduced during reparameterization can enable faster model convergence, but may lead to unstable skeletonization results and deviations from the true medial axis, as illustrated in Fig. \ref{fig_skel}, especially in scenarios with complex geometric structures or topological sensitivity. Finally, the STE mechanism suffers from a mismatch between the gradient used for parameter updates and the true gradient during backpropagation, which may lead to training instability and potentially cause the optimization path to deviate from the desired direction. 

Therefore, we will  propose a method capable of directly computing simple points on continuous-valued images. By iteratively removing these simple points, we obtain a topology-preserving skeletonization algorithm that can be optimized via gradient-based methods. This approach not only achieves higher computational efficiency but also eliminates the need for additional hyperparameter tuning across different datasets. Furthermore, it avoids introducing extraneous noise, as is typical with reparameterization tricks, thereby ensuring the preservation of topological integrity.

\section{The Proposed Method}\label{sec_method}
In this section, we derive the mathematical conditions for computing simple points on continuous-valued images and explain how this theory can be applied to skeleton extraction. Subsequently, we propose a topology-preserving variational model that can be arbitrarily integrated into deep learning based semantic segmentation architectures.

\subsection{Continuous-tone Simple Points}

In the process of determining simple points in a binary image, the essence lies in counting the number of connected components in the foreground and background. This counting operation is non-differentiable. Additionally, we need to define connectivity for continuous-valued images. To address these issues, inspired by the Zhang-Suen thinning algorithm \cite{zhangskel}, we design a method to calculate simple points on continuous-valued images. First, we identify a type of non-simple case, which is termed the non-boundary point.

\begin{definition}[Non-boundary Point]
For $\mathbb{X} \subset \Omega$ and $\mathbb{X}^{c} = \Omega \setminus \mathbb{X}$ represent the foreground and the background, respectively. A point $x \in \Omega$ is called a non-boundary point if $\mathring{\mathbb{N}}_8(x) \cap \mathbb{X} = \emptyset$ or $\mathring{\mathbb{N}}_8(x) \cap \mathbb{X}^{c} = \emptyset.$ 
\end{definition}


Then, we refined the criteria for determining simple points.

\begin{theorem}\label{thm_t4}
    For $\mathbb{X} \subset \Omega$ and $\mathbb{X}^{c} = \Omega \setminus \mathbb{X}$, if the point $x\in \Omega$ is not a non-boundary point, then conditions $T_4(x,\mathbb{X}^{c})=1$ and $T_8(x, \mathbb{X})=1$ in Def. \ref{def_sp} are equivalent to $T_4(x,\mathbb{X}^{c})=1$ alone. 
\end{theorem}

The proof of Theorem \ref{thm_t4} is provided in supplementary material I. To facilitate the calculation of $T_4(x,\mathbb{X}^{c})$, we define a concept that we term the cyclic gradient.
\begin{definition}[Cyclic Gradient Vector]
    Given an image $u(x)$ and the 8-connected neighborhood of a point $x \in \Omega$:
\begin{equation*}
\begin{tikzpicture}[>=Stealth,   
    every node/.style={inner sep=1.5pt},
    inner/.style={->, line width=0.7pt, gray!100},
]
\matrix (m) [matrix of math nodes,
    nodes in empty cells,
    row sep=0.6cm,      
    column sep=0.6cm,   
    left delimiter=(\quad, right delimiter=).
] {
x_1 & x_2 & x_3 \quad \\
x_8 & \bm{x}   & x_4 \quad \\
x_7 & x_6 & x_5 \quad\\
};

\draw[inner, bend left=25] (m-1-1.north) to (m-1-2.north);
\draw[inner, bend left=25] (m-1-2.north) to (m-1-3.north);

\draw[inner, bend left=25] ([xshift=-8pt]m-1-3.east) to ([xshift=-8pt]m-2-3.east);
\draw[inner, bend left=25] ([xshift=-8pt]m-2-3.east) to ([xshift=-8pt]m-3-3.east);

\draw[inner, bend left=25] (m-3-3.south) to (m-3-2.south);
\draw[inner, bend left=25] (m-3-2.south) to (m-3-1.south);

\draw[inner, bend left=25] (m-3-1.west) to (m-2-1.west);
\draw[inner, bend left=25] (m-2-1.west) to (m-1-1.west);

\end{tikzpicture}
\end{equation*}
Then $\mathbb{C}(x) = \{x_1,x_2,x_3,x_4,x_5,x_6,x_7,x_8,x_1\}$ is defined as the cyclic neighborhood of $x$ and the cyclic gradient vector of $u(x)$, denoted by $\nabla_{\mathbb{C}} u(x) = (u(x_1)-u(x_2),u(x_2)-u(x_3),\cdots,u(x_7)-u(x_8), u(x_8)-u(x_1))^\top$.
\end{definition}

In binary image processing, the crossing number is a measure used to characterize the local topology of the $3 \times 3$ neighborhood of a pixel excluding the center, which is equivalent to computing the $\ell_0$ norm of the cyclic gradient vector. 
\begin{definition}[Crossing Number]\label{def_cn}
    The crossing number $\mathcal{C}(u)[x]$ of $x \in \Omega$ is defined as the number of transitions of the pixel value $u(x)$ between the background and the foreground in the cyclic neighborhood $\mathbb{C}(x)$:
    \begin{align*}
        \mathcal{C}(u)[x] = \sum_{i=1}^{8} \delta(|u(x_{i+1}) - u(x_i)|) = || \nabla_{\mathbb{C}} u(x) ||_0,
    \end{align*}
    where $x_9 = x_1$, $u(x_i) \in \{0,1\}$, 
    \begin{align} \label{eq_delta}
        \delta(z)=\begin{cases}
         1, &  z=1, \\
         0, &  z=0.
        \end{cases}
    \end{align}
    Equivalently, it counts the number of times the pattern changes from $0 \to 1$ or $1 \to 0$ and this operation is linear.
\end{definition}

Under the digital topology convention of 8-connectivity for the foreground and 4-connectivity for the background, if $x$ is a non-boundary point, $\mathcal{C}(u)[x]=0$. For other cases, the crossing number is exactly twice the number of 4-connected components of $\mathbb{N}^{1}_8(x,\mathbb{X}^{c})$. Similarly, considering the geodesic neighborhood $\mathbb{N}_4^2(x,\mathbb{X}^{c})$, its distinction from $\mathbb{N}_8^1(x,\mathbb{X}^{c})$ lies in the fact that if $x_{2i},x_{2i+2} \in \mathbb{X}$, where $i = 0,1,\cdots,4, x_9=x_1 ,x_{10}=x_2$, then $x_{2i+1}$ will be considered a foreground point and will not participate in the calculation of the 4-connectivity number (as illustrated in the example in (d) of Fig. \ref{fig_exsp}). Therefore, in this case, if $x$ is not a non-boundary point, 
\begin{align*}
    T_4(x,\mathbb{X}^{c}) &= \frac{1}{2} \mathcal{C}(u_m)[x] \\
    &= \frac{1}{2} \sum_{i=1}^{8} \delta(|u_m(x_{i+1}) - u_m(x_i)|),
\end{align*}
where 
    \begin{align*}
        u_m(x) &= 1 - [m(x) \cdot (1 - u(x))], \\
        m(x) &= \begin{cases}
         1, &  x \in \mathbb{N}_4^2(x,\mathbb{X}^{c}), \\
         0, &  x \in \mathring{\mathbb{N}}_8(x) \setminus \mathbb{N}_4^2(x,\mathbb{X}^{c}).
        \end{cases}
    \end{align*}
Therefore, based on Theorem \ref{thm_t4}, the crossing number can be utilized to determine whether a point is a simple point.

\begin{theorem}
     For $\mathbb{X} \subset \Omega$ and $\mathbb{X}^{c} = \Omega \setminus \mathbb{X}$, the point $x\in \Omega$ is a simple point if and only if $\mathcal{C}(u_m)[x] = 2$. 
\end{theorem}

Next, we extend the binary image case to continuous-valued images. First, we need to define the connectivity of the continuous-valued image.

\begin{definition}[The connectivity of the continuous-valued image] \label{def_conn}
     Given a continuous-valued image $u$, pixels $x,y \in \Omega$ and a threshold $\tau \geq 0$, pixels $x$ and $y$ are said to be $\tau$-connected under $n$-connectivity if the intensity values satisfy $|u(x) - u(y)| \le \tau$ when $y \in \mathbb{N}_n(x)$.
\end{definition}

Since the transition between foreground and background implies a change in connectivity, using Def. \ref{def_conn}, Eq. \eqref{eq_delta} is equivalent to:
\begin{align}\label{eq_delta_tau}
\delta_{\tau}(x)=\begin{cases}
 0, &  x \le \tau, \\
 1, &  \text{else}.
\end{cases}
\end{align}
To enable the application of gradient-based optimization algorithms (such as backpropagation), we employ the sigmoid function to construct a smooth approximation of Eq. \eqref{eq_delta_tau}, leading to the definition of the smooth crossing number.
\begin{definition}[Smooth Crossing Number]
    The smooth crossing number $\mathcal{C}_{\tau}(u)[x]$ of $x \in \Omega$ is defined in the cyclic neighborhood $\mathbb{C}(x)$:
    \begin{equation*}
        \mathcal{C}_{\alpha,\tau}(u)[x] = \sum_{i=1}^{8}  \delta_{\alpha,\tau}(|u(x_{i+1}) - u(x_i)|) \approx || \nabla_{\mathbb{C}} u(x) ||_0,
    \end{equation*}
    where $x_9 = x_1$, $u(x_i) \in [0,1]$, $\alpha > 0$, $\tau \ge 0$ and 
    \begin{align*}
        \delta_{\alpha,\tau}(x)=\frac{1}{1+e^{-\alpha(x-\tau)}}.
    \end{align*}
\end{definition}

Subsequently, we will define a topological detection operator to identify which points are simple points.

\begin{definition}[Topological Detection Operator]
    Given $x \in \Omega$, the Gaussian function is employed to smoothly approximate the indicator function for simple points, resulting in the topological detection operator: 
    \begin{align}\label{eq_W}
        \mathcal{W}_{\alpha,\tau,\sigma}(u)[x] = e^{-\frac{(\frac{1}{2}\mathcal{C}_{\alpha,\tau}(u_m)[x]-1)^2}{2\sigma^2}},
    \end{align}
    where $\mathcal{C}_{\alpha,\tau}(u_m)[x]$ denotes the smooth crossing number calculated with the threshold $\tau$. The corresponding non-smoothed indicator function is
    \begin{align*}
        \mathcal{W}(u)[x]=\begin{cases}
         1, &  \frac{1}{2}\mathcal{C}_{\alpha,\tau}(u_m)[x]=1, \\
         0, &  \text{otherwise},
        \end{cases}
    \end{align*}
    which equals 1 only at simple points. 
\end{definition}

The parameter $\sigma > 0$ controls the degree of smoothing: as $\sigma \to 0$, $\mathcal{W}_{\alpha,\tau,\sigma}(u)[x]$ converges point-wise to the indicator function $\mathcal{W}(u)[x]$. This smooth approximation makes the operator differentiable, facilitating its use in gradient based optimization frameworks. By utilizing the simple point based topological detection function, we can constrain pixel-value changes to occur only at simple points while keeping values at non-simple points unchanged, thereby ensuring that the topological properties of the image remain invariant.

\subsection{Topology Preserving Skeleton Extraction Method} \label{sec_skel}

Using the method for determining continuous-tone simple points introduced in the previous section, we can directly compute the skeleton features of continuous-valued images. In digital images, the most common skeletonization algorithm iteratively removes simple points until only the skeleton remains \cite{skel_survey, 4766974, BERTRAND1996115}. Our skeletonization algorithm follows the same paradigm (see CSPS of Fig. \ref{fig_all}). However, unlike these methods, our algorithm is optimized via true gradient-based methods while ensuring topology preservation. This approach not only achieves higher computational efficiency but also avoids the gradient bias introduced by the STE mechanism. Furthermore, it directly computes simple points on continuous-valued images, circumventing the extraneous noise typically associated with reparameterization techniques, thereby ensuring the preservation of topological integrity.

Furthermore, for image processing tasks — especially in contemporary data-driven approaches — determining whether each individual point is simple in a point-wise manner is highly inefficient. Simultaneously removing all simple points at once is also infeasible because removing adjacent points concurrently may alter the topological structure of the object, even if each point is individually simple. Therefore, we adopt a sub-iteration scheme based on four identical sub-fields, following a method similar to that in \cite{BERTRAND1995}:
\begin{align} \label{eq_order}
    \begin{cases}
    \mathbb{M}_1(\bm x) = \neg (x_1 \; \mathrm{mod} \; 2) \wedge \neg (x_2 \; \mathrm{mod} \; 2), \\
     \mathbb{M}_2(\bm x) = (x_1 \; \mathrm{mod} \; 2) \wedge \neg (x_2 \; \mathrm{mod} \; 2), \\
     \mathbb{M}_3(\bm x) = \neg (x_1 \; \mathrm{mod} \; 2) \wedge (x_2 \; \mathrm{mod} \; 2), \\
     \mathbb{M}_4(\bm x) = (x_1 \; \mathrm{mod} \; 2) \wedge (x_2 \; \mathrm{mod} \; 2), 
    \end{cases}
\end{align}
where $\bm x = (x_1,x_2) \in \Omega$. Performing parallel processing in sequential order $\mathbb{M}_1(\bm x),\mathbb{M}_2(\bm x),\mathbb{M}_3(\bm x),\mathbb{M}_4(\bm x)$ on the image not only significantly improves computational efficiency but also reduces spurious branches on the skeleton while preserving topological invariance.

Moreover, retaining only non-simple points would yield a topological skeleton that compromises the geometric shape of the image. For instance, a solid object without any holes or cavities would be reduced to a single point. This issue can be addressed by simultaneously preserving the so-called endpoints:
\begin{definition}[Endpoint]
    For $x \in \Omega$, $x$ is an endpoint if
    and only if
        $\# \mathring{\mathbb{N}}_n(x) \le 1$,
which is equivalent to computing:
\begin{equation*}
    \mathcal{P}(u)[x] = \min(\max((\sum_{y \in \mathring{\mathbb{N}}_n(x)} \delta_{0.5}(u(y))) -1,0),1),
\end{equation*}
where $n=4$ or $8$, and $\delta_{0.5}$ is defined at Eq. \eqref{eq_delta_tau}.
\end{definition}

When $\mathcal{P}(u)[x] = 1$, it means that $x$ is not an endpoint and can be changed. Then, our designed Continuous-tone Simple Points Skeleton (CSPS) algorithm can be formally described in Algorithm \ref{alg_CSPS}.
\begin{algorithm}[H]
    \caption{Continuous-tone Simple Points
Skeleton (CSPS)}
    \begin{algorithmic}
        \STATE \textbf{Input:} Image $u(x) \in [0,1]$, number of iterations $T$, hyper-parameters $\alpha, \tau, \sigma$.
        \STATE \textbf{Output:} Skeleton $\mathcal{S}(u)$.
        \STATE \textbf{Initializtion:} $\mathcal{S}(u) \leftarrow u.$
        \STATE \textbf{For} $t=0,1,2,\dots,T:$
        \STATE \hspace{0.5cm} $\bullet$ Endpoint determination: 
        \begin{align*}
            \mathcal{P}(u)[x] =\begin{cases}
              0, & x \text{ is endpoint}, \\
              1, & \text{else}.
            \end{cases}
        \end{align*}
        \STATE \hspace{0.5cm} \textbf{For} $i=1,2,3,4:$
        \STATE \hspace{1.0cm} $\bullet$ Simple point determination: 
        $$ \mathcal{W}_{\alpha,\tau,\sigma}(u)[y],\quad y \in \mathbb{M}_i(x). $$
        \STATE \hspace{1.0cm} $\bullet$ Skeletonization:
        $$\mathcal{S}(u) \leftarrow (1 -  \mathcal{P}(u) \cdot \mathcal{W}_{\alpha,\tau,\sigma}(u)) \cdot \mathcal{S}(u).$$
        \STATE \hspace{0.5cm} \textbf{end}
        \STATE \textbf{end}
    \end{algorithmic}
    \label{alg_CSPS}
\end{algorithm}

\subsection{Topology Loss based on Continuous-tone Simple Points} \label{sec_loss}

Based on the skeletons extracted via continuous-tone simple points, we design the loss function following the same strategy as in \cite{cldice}:
\begin{align*}
    T_{prec}(u,g) &= \frac{\sum_{x \in \Omega} \mathcal{S}(u)[x] \cdot g(x)}{\sum_{x \in \Omega} \mathcal{S}(u)[x]}, \\
    T_{sens}(u,g) &= \frac{\sum_{x \in \Omega} \mathcal{S}(g)[x] \cdot u(x)}{\sum_{x \in \Omega} \mathcal{S}(g)[x]}, \\
    \mathcal{L}_{CSP}(u,g) &= 1 - \frac{2 \cdot T_{prec}(u,g) \cdot T_{sens}(u,g)}{T_{prec}(u,g) + T_{sens}(u,g)},
\end{align*}
where $g$ and $u$ denote ground truth and network prediction, respectively, and $\mathcal{S}(\cdot)$ denotes the skeleton extraction operation, i.e., the topologically non-removable points remaining after sequentially removing all simple points. $T_{prec}, T_{sens}$ denote the precision and recall of simple points in the predicted target object, respectively. Subsequently, we add the CSP loss as a regularization term to the binary-cross-entropy loss function to obtain the final loss:
\begin{align*}
    \mathcal{L}(u,g) = \mathcal{L}_{BCE}(u,g) + \lambda \: \mathcal{L}_{CSP}(u,g).
\end{align*}
Essentially, this loss increases the weight of pixels corresponding to non-simple points in the overall image, encouraging the model to focus more on pixels that encode topological properties, thereby enhancing its ability to perceive and represent the global topological structure of the image.

\subsection{General Topology Preserving for Segmentation Network Architecture} \label{sec_model}

\begin{figure*}[!t]
\centering
\includegraphics[width=7.0in]{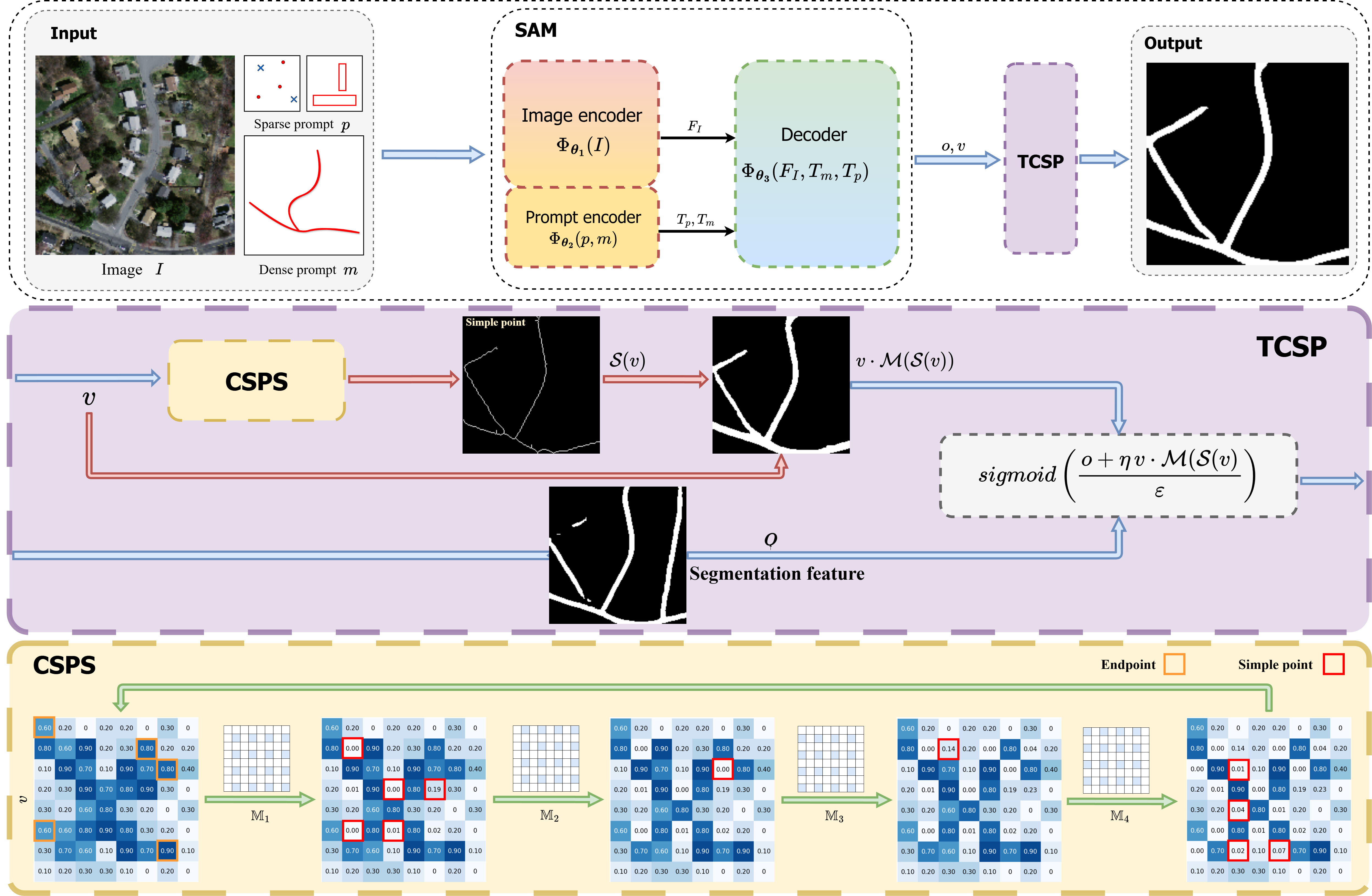}
\caption{Overview of proposed TCSP-SAM network $\mathcal{N}_{\bm \theta}$ defined in Eq. \eqref{eq_tspsam}. TCSP denotes the topology-preserving variational model \eqref{eq_spstd}, and CSPS corresponds to the skeleton extraction algorithm \ref{alg_CSPS}.}
\label{fig_all}
\end{figure*}

Since improving the loss function alone cannot guaranty the enforcement of the desired constraints during the inference stage in modern learning methods, we design a topology-preserving variational model that can be integrated with any semantic segmentation network equipped with softmax or sigmoid activation functions. 

Semantic segmentation networks typically consist of three components: an encoder, a decoder, and a classifier. The image encoder maps the input image to a high-dimensional embedding space, after which the decoder integrates the features extracted by the encoder. Finally, the classifier assigns each pixel to its corresponding semantic class. We now present the mathematical formulation of the SAM \cite{sam}, which, compared to other models, incorporates additional prompt features:
\begin{equation*}
    \begin{cases}
        F_I = \Phi_{\theta_1}(I),\, T_p, T_m = \Phi_{\theta_2}(p, m), \\
        o = \Phi_{\theta_3}(F_I, T_m, T_p), \\
        u^{\ast} = \delta_0(o),
    \end{cases}
\end{equation*}
where $m, p$ are dense and sparse prompts, respectively, $F_I, T_m, T_p$ indicate the image feature, the dense embedding and the sparse embedding, and $o$ is the segmentation feature extracted by SAM. $\delta_0$ is a classifier defined in Eq. \eqref{eq_delta_tau}, which is equivalent to the minimization problem:
\begin{align*}
    u^{\ast} = \delta_0(o) &= \arg\min\limits_{u\in[0,1]}\{ \langle -o, u\rangle \} \\
    &= \arg\min\limits_{u\in[0,1]}\{ \sum_{x \in \Omega} -o(x)\cdot u(x) \}.
\end{align*}

Then, only improving the loss function does not guaranty the preservation of topological information during inference. To address this issue, building on previous work by our group in \cite{std}, we integrate simple point based topological information into a data-driven model, named the Topological Continuous-tone Simple Points (TCSP) model:
\begin{equation}
    \min\limits_{u \in [0,1]} \mathcal{E}(u, o, v) := \min\limits_{u \in [0,1]} \{ \langle-o,u\rangle+ \varepsilon \mathcal{H}(u) + \eta \mathcal{T}(u,v) \}, \label{eq_spstd}
\end{equation}
where
\begin{align*}
    &\mathcal{H}(u) = \langle u,\ln u\rangle  +\langle 1-u, \ln(1-u) \rangle, \\
    &\mathcal{T}(u,v) = \langle 1-u, v \cdot \mathcal{M}(\mathcal{S}(v)) \rangle, \\
    &\mathcal{M}(\mathcal{S}(v))[x] = \max_{y \in \mathbb{B}(x,r)}\mathcal{S}(v)[y].
\end{align*}
Here, $u,g$ is the predicted segmentation result and the ground-truth, $o,v$ denote the segment feature and the feature for identifying topologically non-simple points $\mathcal{S}(v)$, respectively, both extracted from the image by neural networks. $\mathcal{M}$ is the dilation operator in morphology and $\mathbb{B}(x,r)$ denotes the neighborhood of the structuring element centered on $x$ with radius $r$. Additionally, in model \eqref{eq_spstd}, $\mathcal{H}(u)$ represents an entropy regularization term that smooths the segmentation result and promotes stability during the iterative optimization process of the network. 

\begin{figure}[!t]
\centering
\includegraphics[width=3.45in]{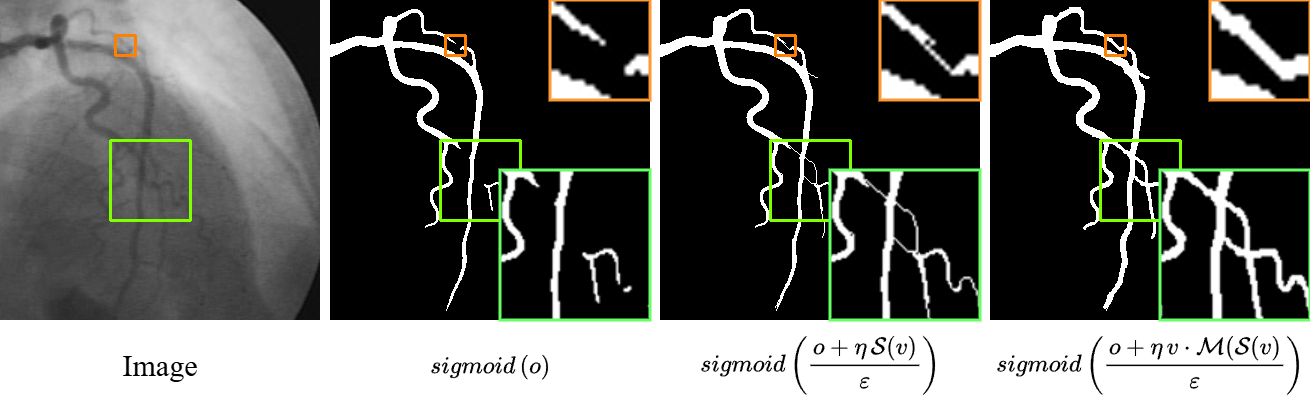}
\caption{Visualization of results with the regularization term $\mathcal{T}(u,v)$ and the connectivity extraction operator incorporated into $\mathcal{T}(u,v)$.}
\label{fig_connex}
\end{figure}

The term $\mathcal{T}(u,v)$ is a topological simple point regularization term. $\mathcal{S}(v)$ denote the set of points whose value change would alter the topological properties. According to the order $\mathbb{M}_1(\bm x),\cdots, \mathbb{M}_4(\bm x)$ in Eq. \eqref{eq_order} in which we determine the simple points, $\mathcal{S}(v)$ appears as a skeleton that shares the same topological structure as $v$. However, this would compromise its geometric structure. Therefore, we employ a morphological connected component extraction algorithm $v \cdot \mathcal{M}(\mathcal{S}(v))$ to restore it, as shown in Fig \ref{fig_connex}. Since $1-u$ denotes the background, $\mathcal{T}(u,v)$ penalizes the probability that topologically non-removable points lie in the background of the segmentation result $u$, thereby ensuring that the non-simple points $\mathcal{S}(v)$ always remain in the foreground, which guaranties topological consistency. In addition, the problem \eqref{eq_spstd} is strongly convex with respect to $u$ and has a closed-form solution $u^{\ast}$ (see supplementary material II for details):
\begin{equation*}
    \begin{split}
        u^{\ast} &= sigmoid\left(\frac{o + \eta \,  v \cdot \mathcal{M}(\mathcal{S}(v))}{\varepsilon}\right) \\
        &=\left(1+exp(-\frac{o + \eta \,  v \cdot \mathcal{M}(\mathcal{S}(v))}{\varepsilon})\right)^{-1}.
    \end{split}
\end{equation*}

Therefore, TCSP-SAM can be represented as a network $\mathcal{N}_\theta$ parameterized by $\theta = (\theta_1, \theta_2, \theta_3)$ that maps an image $I$ and prompt information $p, m$ to a pair $(u^*, v)$, where $u^*$ is the segmentation result that preserves topological information and $v$ is the auxiliary variable for the topology constraint. It can be written as:
\begin{align}
    \label{eq_tspsam}
    \mathcal{N}_{\bm\theta}(I,p,m)=\begin{cases}
    F_I = \Phi_{\theta_1}(I),\, T_p, T_m = \Phi_{\theta_2}(p, m), \\
    o,v = \Phi_{\theta_3}(F_I,T_m, T_p), \\
    u^{\ast} = \arg \min\limits_{u \in [0,1]} \mathcal{E}(u, o, v),
    \end{cases}
\end{align}
where $\mathcal{E}(u)$ is the TSCP model \eqref{eq_spstd} and the architecture of $\mathcal{N}_\theta$ is shown in Fig. \ref{fig_all}. Thus, the training of the proposed model can be expressed as:
\begin{align*}
    \bm\theta^*=\arg\min_{\bm\theta} \left\{ \mathcal{L}_{BCE}( u^*, g) + \lambda \: \mathcal{L}_{CSP}(v,g) \right\},
\end{align*}
where $(u^*, v) = \mathcal{N}_\theta(I, p, m)$.

The proposed TCSP-SAM integrates differentiable topology preservation directly into the segmentation framework. By learning to identify topologically non‑removable points $\mathcal{S}(v)$ and enforcing the variational constraint $\mathcal{T}$, it effectively maintains global topological accuracy by keeping non-simple points always in the foreground and leverages true gradients for optimization. Subsequently, we will evaluate the performance of the continuous‑tone simple point (CSP) method through numerical experiments.

\section{Experiments}\label{sec_experiment}
In this section, we validate the capability of the continuous-tone simple point (CSP) method to preserve topology in skeleton extraction and semantic segmentation. In addition, we conducted comparative experiments with the most relevant existing methods, including morphology-based methods \cite{cldice} and a method that computes the gradient of simple points using STE and reparameterization (denoted SRSP) \cite{spskel}. In all experiments, the parameters of Eq. \eqref{eq_W} are set to $\alpha = 16$, $\sigma=0.2$ and $\tau=0.5$. All computations are performed using Python 3.11.11 and PyTorch 2.4.0 on an NVIDIA 4090 GPU. 

\subsection{Datasets and Preprocessing}
Several datasets that contain tubular and reticular structures were utilized in our experiments due to the stringent requirements for geometry and detail.
\begin{itemize}
    \item DRIVE dataset \cite{drive}: Originating from the Dutch diabetic retinopathy screening program, this dataset comprises 40 retinal images along with the corresponding annotations of retinal blood vessels. The images were captured with 8 bits per color channel at a resolution of 565 × 584 pixels. The dataset has been officially partitioned into a training set and a test set, each containing 20 images.
    \item DCA dataset \cite{dca}: Comprising 134 coronary angiography images along with their corresponding ground-truth annotations labeled by expert cardiologists. The entire image database was provided by the Mexican Social Security Institute, UMAET1-León. Each angiographic image is a grayscale image with a resolution of 300 × 300 pixels. The database is divided into two subsets: the training set contains 100 images, and the remaining 34 images constitute the test set.
    \item MASS Dataset \cite{mass}: Massachusetts Road dataset consists of 1171 aerial color images with a resolution of 1500 × 1500 pixels, covering urban, rural and mountainous areas in Massachusetts, USA. We cropped each image into 800 × 800 pixel patches and randomly selected 1,600 patches for the training set, while adopting the official splits for the validation and test sets.
    \item UBW dataset \cite{ubw}: Urinary Bladder Wall dataset consists of 1281 magnetic resonance imaging (MRI) images (512 × 512 pixels), which was challenged at the Third International Symposium on Image Computing and Digital Medicine (ISICDM 2019). Since the bladder is consistently located in the center of the images, we applied a center crop of 320 × 320 pixels to each image. The dataset was then partitioned into training, validation and test sets based on slices from different patients, with a split ratio of $904 : 185 : 192$.
\end{itemize}
During training, to ensure experimental fairness, we applied identical preprocessing steps to all models and methods, including random cropping, horizontal and vertical flipping, and normalization. During testing, only normalization was applied.

\subsection{Evaluation Metrics}
For evaluation metrics, we compare the performance of various experimental setups using three types: overlap-based, boundary-based, and topology-based.
\begin{itemize}
    \item Overlap-based: Recall, Intersection Over Union (IoU) and Dice score coefficient (Dice).
    \item Boundary-based: 95$\%$ Hausdorff distance (HD95) \cite{hd95} and Average Symmetric Surface Distance (ASSD) \cite{assd}.
    \item Topology-based: cl-Dice \cite{cldice}, 0- and 1-Betti numbers error($\beta_0$ and $\beta_1$) \cite{gudhi}, and Euler Characteristics error ($\chi_{error}$) \cite{skimage}.
\end{itemize}

\subsection{Validate Effectiveness in Skeleton Extraction}

\begin{figure*}[!t]
\centering
\includegraphics[width=6.5in]{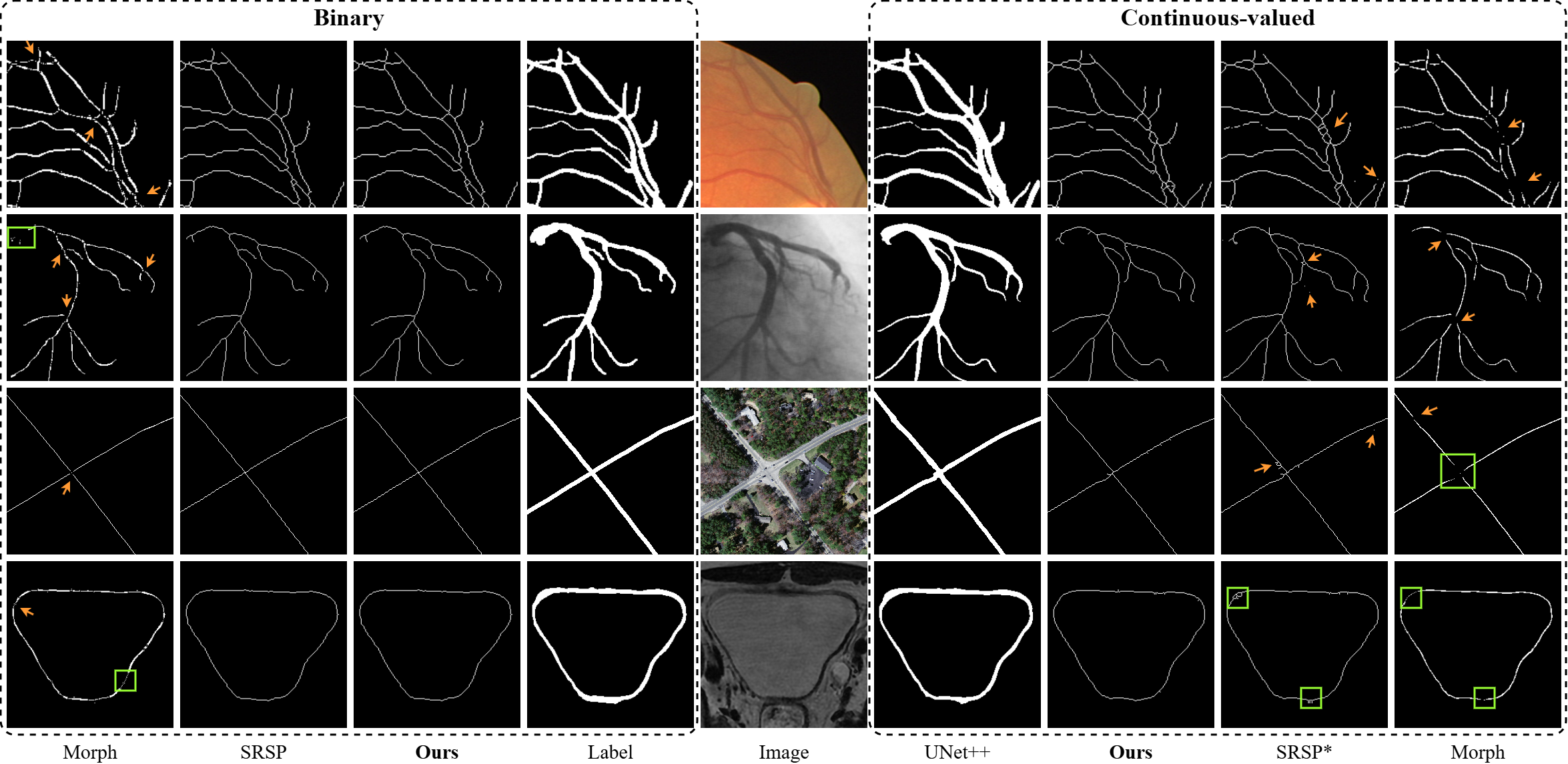}
\caption{Visualization of skeleton extraction results from binary and continuous-valued images using three methods. SRSP* denotes the addition of logical noise with intensity 0.1.}
\label{fig_skel}
\end{figure*}

\begin{table*}[htbp]
\scriptsize
\centering
\caption{Quantitative comparison of the topographic accuracy and run time of several skeletonization algorithms for binary image.}
\label{tab:comparison}
\begin{tabular}{lllccc|c|lllccc}
\toprule
Image & Dataset & Skeleton & $\beta_0 \downarrow $ & $\beta_1 \downarrow $ & $\chi_{error} \downarrow $ & Run time [ms] & Image & Dataset & Skeleton & $\beta_0 \downarrow $ & $\beta_1 \downarrow $ & $\chi_{error} \downarrow $ \\
\midrule
\multirow{12}{*}{Binary} & \multirow{3}{*}{DCA} 
& Morph \cite{cldice} & $133.65$ & $0.79$ & 134.44 & $4.09$ & \multirow{12}{*}{Continuous-valued} & \multirow{3}{*}{DCA} & Morph \cite{cldice} & 18.00 & 0.88 & 18.88 \\
& & SRSP \cite{spskel} & $0.00$ & $0.00$ & 0.00 & $144.52$ & & & SRSP* \cite{spskel} & 2.21 & 0.68 & 2.00 \\
& & Ours & $0.00$ & $0.00$ & 0.00 & $33.40$ & & & Ours & 0.00 & 0.00 & 0.00 \\
\cmidrule(lr){2-6} \cmidrule(lr){7-7} \cmidrule(lr){9-13}
& \multirow{3}{*}{DRIVE} 
& Morph \cite{cldice} & $768.80 $ & $53.95$ & 822.75 & $5.14$ & & \multirow{3}{*}{DRIVE} & Morph \cite{cldice} & 212.25 & 27.75 & 240.00 \\
& & SRSP \cite{spskel} & $0.00$ & $0.00$ & 0.00 & $74.27$ & & & SRSP* \cite{spskel} & 25.70 & 1.55 & 24.65 \\
& & Ours & $0.00$ & $0.00$ & 0.00 & $16.82$ & & & Ours & 0.00 & 0.00 & 0.00 \\
\cmidrule(lr){2-6} \cmidrule(lr){7-7} \cmidrule(lr){9-13}
& \multirow{3}{*}{MASS} 
& Morph \cite{cldice} & $235.07$ & $9.84$ & 244.91 & $4.13$ & & \multirow{3}{*}{MASS} & Morph \cite{cldice} & 70.22 & 4.26 & 74.48 \\
& & SRSP \cite{spskel} & $0.00$ & $0.00$ & 0.00 & $104.13$ & & & SRSP* \cite{spskel} & 30.09 & 17.81 & 15.14 \\
& & Ours & $0.00$ & $0.00$ & 0.00 & $17.15$ & & & Ours & 0.00 & 0.00 & 0.00 \\
\cmidrule(lr){2-6} \cmidrule(lr){7-7} \cmidrule(lr){9-13}
& \multirow{3}{*}{UBW} 
& Morph \cite{cldice} & $47.06$ & $1.01$ & 48.07 & $4.79$ & & \multirow{3}{*}{UBW} & Morph \cite{cldice} & 5.46 & 0.97 & 6.43 \\
& & SRSP \cite{spskel} & $0.00$ & $0.00$ & 0.00 & $144.62$ & & & SRSP* \cite{spskel} & 1.10 & 1.57 & 1.46 \\
& & Ours & $0.00$ & $0.00$ & 0.00 & $33.39$ & & & Ours & 0.00 & 0.00 & 0.00 \\
\bottomrule
\end{tabular}\\
\smallskip
Note: SRSP* denotes that logical noise with intensity 0.1 was added, as in \cite{spskel}. In the absence of noise, the three topological metrics mentioned above are all $\boldsymbol{0}$, indicating that no topological changes occur. However, as noted in \cite{spskel}, omitting noise can inhibit learning.
\label{tbl_skel}
\end{table*}

This section validates the effectiveness of our method in skeleton extraction tasks corresponding to Sec. \ref{sec_skel}. We apply the proposed method to extract skeletons from both binary images and prediction maps of segmentation networks, and compare it with the two most relevant approaches, namely Morph \cite{cldice} and SRSP \cite{spskel}. Quantitative and qualitative results are presented in Table \ref{tbl_skel} and Fig. \ref{fig_skel}, respectively. 

As shown in Table \ref{tbl_skel}, our method significantly improves computational efficiency compared to SRSP while preserving topological consistency, for binary images and predicted probability maps of neural networks. This is crucial for data-driven approaches. Additionally, we also evaluated the SRSP method with logical noise (denoted as SRSP*). Although the original paper notes that adding appropriate noise can accelerate convergence during training, doing so tends to disrupt the original topological structure. Moreover, the SRSP method does not generalize well across different datasets in terms of parameter settings.

\subsection{Validation of Effectiveness for loss-based Semantic Segmentation}

In loss-based manner, we incorporate our method into a baseline segmentation framework and compare the segmentation performance with four types of loss functions in each dataset. The first is training with the baseline loss only, the second adds a morphology-based loss \cite{cldice} to the baseline, the third adds an SRSP loss \cite{spskel} and the fourth adds our proposed CSP loss. The same training strategy is adopted for all four loss configurations. We aim to examine whether CSP loss can maintain overall topological similarity, regardless of the specific network architecture. The specific experimental details are presented below.
\begin{itemize}
    \item Benchmark Segmentation Frameworks: We select several open-source classic segmentation frameworks, including DeepLabV3+ \cite{deeplabv3+}, SegFormer \cite{segformer}, UNet++ \cite{unet++}, and SAM2 \cite{sam2}. The pre-training encoder weights of UNet++, Deeplabv3+ are ResNet101, SegFormer is MiT-B5 and SAM2 is sam2.1-hiera-base-plus.
    \item Training Setups: We train these models using the AdamW (Adaptive Moment Estimation with Decoupled Weight Decay) optimizer \cite{adamw} under the same learning strategy. For SAM2, the learning rate is set to 1e-4 with a batch size of 2, using 30 epochs for the MASS and UBW datasets and 50 epochs for DRIVE and DCA. For the remaining models, the learning rate is 3e-4 with a batch size of 4, and the number of epochs is 100 for MASS and UBW, and 200 for DRIVE and DCA. Additionally, due to the complexity of tubular structures, we did not use prompts in SAM2 and performed full fine-tuning. Finally, the regularization parameter of the loss function is set to $\lambda = 0.001$.
\end{itemize}

\begin{figure}[!t]
\centering
\includegraphics[width=3.4in]{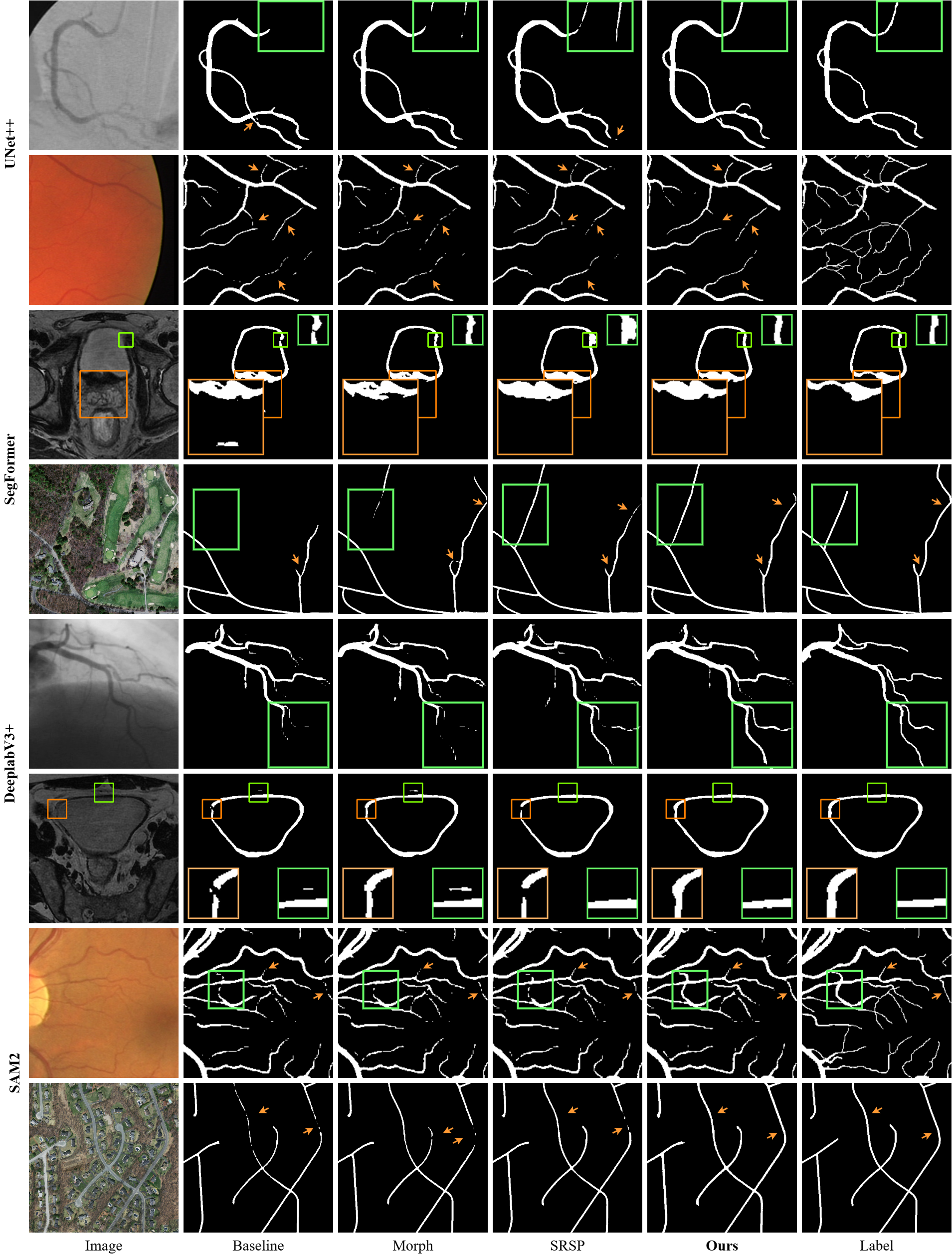}
\caption{Segmentation results of different loss functions. “Baseline” denotes training with cross-entropy loss only; “Morph” and “SRSP” correspond to the methods proposed in \cite{cldice} and \cite{spskel}, respectively; “Ours” denotes the proposed CSP method.}
\label{fig_loss}
\end{figure}

Then, evaluation metrics are computed on the test set and the models trained with the four loss functions are compared and analyzed. Table \ref{tab_loss} presents the segmentation performance of baseline networks trained with different loss functions on the test sets of various datasets. Compared to baseline cross-entropy loss, incorporating the morphology-based method (Morph) \cite{cldice} leads to improvements in overlap-based and boundary-based metrics, but may adversely affect topology-related metrics, as morphological methods inherently lack topological consistency constraints. In addition, both the SRSP method \cite{spskel} and our proposed CSP method achieve strong performance on topological metrics. However, due to potential bias in the propagated gradients during backpropagation, SRSP exhibits weaker performance on overlap-based metrics. In contrast, our method achieves strong performance in nearly all comparisons with few exceptions. Fig. \ref{fig_loss} provides a more intuitive visualization, showing that the segmentation results generated by the network trained with the CSP loss are geometrically and topologically closer to the ground truth.

\begin{table*}[htbp]
\scriptsize
\centering
\caption{Test results of baseline networks trained with different loss functions on four datasets..}
\label{tab_loss}
\begin{tabular}{ll|ccc|cc|cccc}
\toprule
Dataset & model & Recall $\% \uparrow$ & Dice $\% \uparrow$ & IoU $\% \uparrow$ & HD95 $\downarrow$ & ASSD $\downarrow$ & clDice $\% \uparrow$ & $\beta_0 \downarrow $ & $\beta_1 \downarrow $ & $\chi_{error} \downarrow $ \\
\midrule
\multirow{16}{*}{DCA} & Unet++ \cite{unet++} & 80.1003 & 79.8925 & 66.6710 & 15.3704 & 2.5470 & 85.6700 & 6.1765 & 0.6471 & 6.2941 \\
& \, Morph \cite{cldice} & 80.8722 & 80.1767 & 67.0439 & 13.8225 & 2.4584 & 86.1250 & 5.7059 & \textbf{0.5882} & 5.8235 \\
& \, SRSP \cite{spskel} & \textbf{82.1849} & 80.0562 & 66.8712 & 14.8424 & 2.5139 & 85.8441 & 4.5588 & 0.6471 & \textbf{4.5588} \\
& \, Ours & 81.1558 & \textbf{80.4757} & \textbf{67.4463} & \textbf{11.6241} & \textbf{2.2061} & \textbf{86.7664} & \textbf{4.5588} & 0.7353 & 4.7647 \\
\cmidrule(lr){2-2} \cmidrule(lr){3-5} \cmidrule(lr){6-7} \cmidrule(lr){8-11}
 & SegFormer \cite{segformer} & 82.3124 & 78.7082 & 65.0651 & 13.1641 & 2.4328 & 84.7611 & 13.5882 & 0.9118 & 13.8529 \\
 & \, Morph \cite{cldice} & 81.9008 & 78.8133 & 65.2070 & 11.6997 & 2.4258 & 84.9768 & 14.9118 & 0.9412 & 15.0294 \\
& \, SRSP \cite{spskel} & \textbf{83.0597} & 78.8971 & 65.3086 & 13.1609 & 2.4429 & 84.9161 & 11.4118 & \textbf{0.7941} & 11.4412 \\
& \, Ours & 82.9267 & \textbf{78.9505} & \textbf{65.3901} & \textbf{11.6177} & \textbf{2.2997} & \textbf{85.6496} & \textbf{9.7941} & 0.8235 & \textbf{9.6765} \\
\cmidrule(lr){2-2} \cmidrule(lr){3-5} \cmidrule(lr){6-7} \cmidrule(lr){8-11}
 & DeeplabV3+ \cite{deeplabv3+} & 80.2572 & 78.6343 & 64.9701 & 12.3453 & 2.4142 & 83.5669 & 16.5294 & 0.7647 & 16.7059 \\
 & \, Morph \cite{cldice} & 80.0691 & 78.4144 & 64.6572 & 13.5898 & 2.4800 & 83.5397 & 15.7647 & 0.9706 & 15.9118 \\
& \, SRSP \cite{spskel}& 82.3439 & 78.6182 & 64.9013 & 18.0609 & 2.7716 & 83.6854 & 15.5882 & 0.7941 & 15.7353 \\
& \, Ours & \textbf{84.1272} & \textbf{79.0299} & \textbf{65.4782} & \textbf{11.3172} & \textbf{2.3792} & \textbf{85.0708} & \textbf{11.9118} & \textbf{0.6471} & \textbf{11.8529} \\
\cmidrule(lr){2-2} \cmidrule(lr){3-5} \cmidrule(lr){6-7} \cmidrule(lr){8-11}
 & SAM2 \cite{sam2} & 83.0045 & 81.2827 & 68.5940 & \textbf{9.5211} & 1.9925 & 88.0220 & 5.2353 & 0.7941 & 5.0294 \\
 & \, Morph \cite{cldice} & 82.9714 & 81.4950 & 68.8988 & 11.0748 & 2.0742 & 87.9613 & 5.7647 & \textbf{0.7941} & 5.5588 \\
& \, SRSP \cite{spskel} & 83.0479 & 81.5304 & 68.9278 & 9.6276 & 1.9723 & 88.3289 & 5.3235 & 1.0294 & 5.0000 \\
& \, Ours & \textbf{83.1488} & \textbf{81.5520} & \textbf{68.9595} & 9.5617 & \textbf{1.9607} & \textbf{88.3970} & \textbf{5.0000} & 1.0000 & \textbf{4.7059} \\
\midrule
\multirow{16}{*}{DRIVE} & Unet++ \cite{unet++} & 79.7540 & 81.6053 & 68.9702 & 5.1188 & 1.1287 & 81.5738 & 98.7500 & 31.2000 & 129.9500 \\
& \, Morph \cite{cldice} & 80.1331 & 81.6845 & 69.0731 & 5.2022 & 1.1294 & 81.6149 & 108.4000 & 31.0500 & 139.4500 \\
& \, SRSP \cite{spskel} & 79.3733 & 81.5808 & 68.9318 & 5.4414 & 1.1630 & 81.2447 & 102.9500 & 31.6000 & 134.5500 \\
& \, Ours & \textbf{80.5701} & \textbf{81.7572} & \textbf{69.1793} & \textbf{5.0886} & \textbf{1.1202} & \textbf{81.7277} & \textbf{95.0500} & \textbf{30.8000} & \textbf{125.8500} \\
\cmidrule(lr){2-2} \cmidrule(lr){3-5} \cmidrule(lr){6-7} \cmidrule(lr){8-11}
 & SegFormer \cite{segformer} & 76.8746 & 77.0885 & 62.7423 & 7.6307 & 1.5060 & 74.4092 & 182.7000 & 38.0000 & 220.7000 \\
 & \, Morph \cite{cldice} & 76.5792 & 76.9342 & 62.5433 & 7.7575 & 1.5349 & 74.2207 & 182.9500 & 37.4500 & 220.4000 \\
& \, SRSP \cite{spskel} & 76.7368 & 77.1686 & 62.8518 & 7.9064 & 1.5336 & 74.3710 & 181.1000 & 38.7000 & 219.8000 \\
& \, Ours & \textbf{78.7972} & \textbf{77.2581} & \textbf{62.9646} & \textbf{6.3696} & \textbf{1.3995} & \textbf{75.8319} & \textbf{177.8500} & \textbf{36.3500} & \textbf{214.2000} \\
\cmidrule(lr){2-2} \cmidrule(lr){3-5} \cmidrule(lr){6-7} \cmidrule(lr){8-11}
 & DeeplabV3+ \cite{deeplabv3+} & 71.1246 & 75.7510 & 61.0227 & 9.4046 & 1.6647 & 72.1130 & 199.5500 & 42.3000 & 241.8500 \\
 & \, Morph \cite{cldice} & 72.5470 & 76.3447 & 61.8056 & 9.3125 & 1.6724 & 72.9727 & 187.9500 & 41.7000 & 229.6500 \\
& \, SRSP \cite{spskel} & 73.5726 & 76.6021 & 62.1307 & 9.4933 & 1.6954 & 73.4419 & \textbf{173.1000} & 42.2000 & \textbf{215.3000} \\
& \, Ours & \textbf{74.1307} & \textbf{77.1607} & \textbf{62.8503} & \textbf{7.9743} & \textbf{1.5102} & \textbf{73.7537} & 196.3500 & \textbf{40.4500} & 236.8000 \\
\cmidrule(lr){2-2} \cmidrule(lr){3-5} \cmidrule(lr){6-7} \cmidrule(lr){8-11}
 & SAM2 \cite{sam2} & 80.4850 & 82.2150 & 69.8286 & 5.9204 & 1.1868 & 80.9068 & 146.7000 & 34.4000 & 181.1000 \\
 & \, Morph \cite{cldice} & 81.0315 & \textbf{82.3275} & \textbf{69.9882} & 5.3449 & 1.1372 & 81.1931 & 155.8000 & 34.7500 & 190.5500 \\
& \, SRSP \cite{spskel} & 81.9620 & 82.2959 & 69.9425 & 4.6819 & 1.0783 & 82.0062 & \textbf{140.1500} & 33.9000 & \textbf{174.0500} \\
& \, Ours & \textbf{82.6406} & 82.3224 & 69.9762 & \textbf{4.5711} & \textbf{1.0748} & \textbf{82.0546} & 142.3000 & \textbf{32.7000} & 175.0000 \\
\midrule
\multirow{16}{*}{MASS} & Unet++ \cite{unet++} & 67.1599 & 74.4496 & 60.5118 & 49.6324 & 7.7535 & 84.6244 & 8.1311 & 5.7705 & 13.6284 \\
& \, Morph \cite{cldice} & 70.0515 & 75.6715 & 61.8787 & 43.3719 & 6.6911 & 85.7654 & 9.9727 & 5.7705 & 15.4481 \\
& \, SRSP \cite{spskel} & 69.3376 & 75.5546 & 61.6687 & \textbf{39.9515} & \textbf{6.3462} & 85.8352 & 9.0820 & 5.8033 & 14.6557 \\
& \, Ours & \textbf{70.1514} & \textbf{76.1149} & \textbf{62.4333} & 46.0304 & 7.2056 & \textbf{86.1511} & \textbf{6.8689} & \textbf{5.3989} & \textbf{12.0492} \\
\cmidrule(lr){2-2} \cmidrule(lr){3-5} \cmidrule(lr){6-7} \cmidrule(lr){8-11}
 & SegFormer \cite{segformer} & 73.3390 & 76.2458 & 62.5774 & 45.9077 & 7.5675 & 86.2695 & 9.3770 & \textbf{4.6721} & 13.0765 \\
 & \, Morph \cite{cldice} & 70.8895 & 75.8106 & 61.7989 & \textbf{43.2171} & \textbf{6.9383} & 86.4190 & 12.9945 & 5.0929 & 17.0492 \\
& \, SRSP \cite{spskel} & 70.5946 & 75.9534 & 61.9983 & 46.5926 & 7.1547 & 86.7224 & 10.3169 & 5.3005 & 14.6011 \\
& \, Ours & \textbf{74.7700} & \textbf{76.8276} & \textbf{63.1588} & 45.4378 & 7.1151 & \textbf{87.0340} & \textbf{8.6448} & 4.6776 & \textbf{12.3934} \\
\cmidrule(lr){2-2} \cmidrule(lr){3-5} \cmidrule(lr){6-7} \cmidrule(lr){8-11}
 & DeeplabV3+ \cite{deeplabv3+} & 62.5339 & 67.8963 & 52.5567 & 49.4338 & 8.7534 & 78.3398 & 41.6503 & 8.0164 & 49.4699 \\
 & \, Morph \cite{cldice} & 66.6539 & 72.0724 & 57.3001 & 43.9223 & 7.0076 & 82.7340 & 26.3934 & 7.1421 & 33.2623 \\
& \, SRSP \cite{spskel} & 66.7646 & 70.9509 & 56.2137 & 44.5096 & 8.6591 & 81.3826 & 29.3224 & 7.2404 & 36.3115 \\
& \, Ours & \textbf{67.2384} & \textbf{72.4043} & \textbf{57.6715} & \textbf{38.3571} & \textbf{6.3317} & \textbf{83.2265} & \textbf{24.6393} & \textbf{6.7596} & \textbf{31.1913} \\
\cmidrule(lr){2-2} \cmidrule(lr){3-5} \cmidrule(lr){6-7} \cmidrule(lr){8-11}
 & SAM2 \cite{sam2} & 72.0087 & 76.5137 & 62.9601 & 52.0128 & 8.1508 & 86.2637 & 7.0164 & 5.0492 & 10.9290 \\
 & \, Morph \cite{cldice} & 72.6158 & 76.8410 & 63.2434 & 49.9710 & 7.8461 & 86.5739 & 6.9235 & 4.8798 & 10.9071 \\
& \, SRSP \cite{spskel} & 72.7685 & 77.0055 & \textbf{63.5454} & 48.5860 & 7.7376 & 86.6873 & \textbf{6.6667} & 4.7541 & 10.7869 \\
& \, Ours & \textbf{76.7728} & \textbf{77.0591} & 63.3338 & \textbf{45.9222} & \textbf{7.6505} & \textbf{87.5355} & 7.8634 & \textbf{4.0820} & \textbf{10.5464} \\
\midrule
\multirow{16}{*}{UBW} & Unet++ \cite{unet++} & 82.3980 & 81.1828 & 69.7261 & 14.4719 & 2.5113 & 90.3264 & 1.2240 & 0.5573 & 1.5729 \\
& \, Morph \cite{cldice} & 78.8670 & 80.8103 & 69.1719 & 7.4485 & 1.6705 & 90.7572 & 1.7448 & 0.5104 & 2.1302 \\
& \, SRSP \cite{spskel} & 82.0043 & 82.4794 & 71.3960 & 9.0473 & 2.0376 & 91.7098 & 1.1302 & 0.5313 & 1.4635 \\
& \, Ours & \textbf{85.8386} & \textbf{84.0257} & \textbf{73.3847} & \textbf{3.2776} & \textbf{1.0187} & \textbf{94.1610} & \textbf{0.4688} & \textbf{0.4844} & \textbf{0.8385} \\
\cmidrule(lr){2-2} \cmidrule(lr){3-5} \cmidrule(lr){6-7} \cmidrule(lr){8-11}
 & SegFormer \cite{segformer} & 83.3984 & 82.6960 & 71.6117 & 4.7720 & 1.2221 & 93.1050 & 0.5417 & 0.3802 & 0.9010 \\
 & \, Morph \cite{cldice} & 81.8760 & 83.0102 & 72.0867 & \textbf{3.7431} & \textbf{1.0802} & 92.8680 & 0.8646 & 0.4219 & 1.2656 \\
& \, SRSP \cite{spskel} & \textbf{83.5575} & 82.8542 & 71.8212 & 4.2077 & 1.2040 & 93.3857 & \textbf{0.4896} & 0.4010 & \textbf{0.8698} \\
& \, Ours & 83.4072 & \textbf{83.4615} & \textbf{72.5397} & 4.0491 & 1.1287 & \textbf{93.9240} & 0.5156 & \textbf{0.3594} & 0.8750 \\
\cmidrule(lr){2-2} \cmidrule(lr){3-5} \cmidrule(lr){6-7} \cmidrule(lr){8-11}
 & DeeplabV3+ \cite{deeplabv3+} & 83.7570 & 82.3428 & 71.2814 & 5.5656 & 1.4120 & 92.4963 & 0.8594 & 0.4896 & 1.2969 \\
 & \, Morph \cite{cldice} & 83.5183 & 83.0905 & 72.2437 & 4.4085 & 1.1601 & 92.5471 & 1.1094 & 0.4844 & 1.5625 \\
& \, SRSP \cite{spskel} & 83.7396 & 82.8103 & 71.6813 & \textbf{3.8995} & 1.1435 & 92.7868 & 1.1198 & 0.4427 & 1.5313 \\
& \, Ours & \textbf{84.6500} & \textbf{83.7559} & \textbf{73.2205} & 4.1438 & \textbf{1.1203} & \textbf{93.5159} & \textbf{0.5990} & \textbf{0.3750} & \textbf{0.9219} \\
\cmidrule(lr){2-2} \cmidrule(lr){3-5} \cmidrule(lr){6-7} \cmidrule(lr){8-11}
 & SAM2 \cite{sam2} & 83.7872 & 83.7258 & 73.1159 & 4.4944 & 1.3016 & 93.0102 & 0.9740 & 0.4740 & 1.2083 \\
 & \, Morph \cite{cldice} & 84.2121 & 83.5228 & 72.7621 & 3.5893 & 1.0773 & 93.3704 & 0.8958 & 0.4271 & 1.0104 \\
& \, SRSP \cite{spskel} & \textbf{85.5278} & 84.4254 & 74.1732 & 3.5702 & 1.0396 & 93.3909 & \textbf{0.6042} & \textbf{0.3802} & \textbf{0.8594} \\
& \, Ours & 85.1967 & \textbf{84.5309} & \textbf{74.2561} & \textbf{3.5191} & \textbf{1.0265} & \textbf{93.6348} & 0.8698 & 0.4010 & 1.0417 \\
\bottomrule
\end{tabular}
\end{table*}

\subsection{Validation of the Topology-Preserving Variational Model for Semantic Segmentation}

\begin{figure}[!t]
\centering
\includegraphics[width=3.5in]{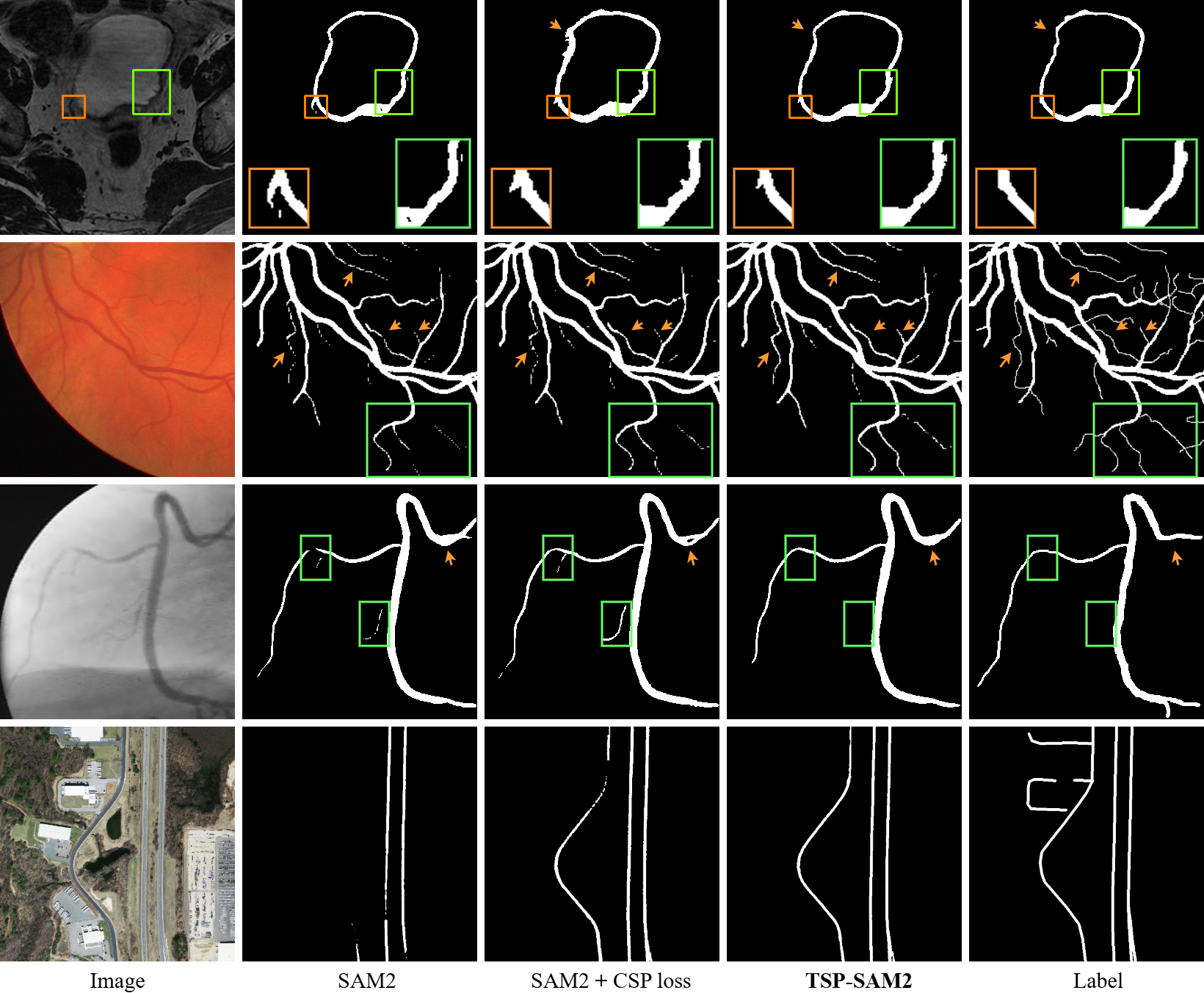}
\caption{Visualization results of the CSP loss and the topology-preserving variational model TCSP-SAM2.}
\label{fig_model}
\end{figure}

Building on SAM2, we construct a topology-preserving segmentation model, TCSP-SAM2, using a variational approach. The TCSP component, in conjunction with the CSPS skeleton extraction algorithm, aims to enhance topological and geometric consistency in semantic segmentation by explicitly preserving the connected components corresponding to topologically non-removable points (i.e., non-simple points) during the segmentation process, as detailed in Sec \ref{sec_model}. We validate its performance by comparing it with the SAM2 trained with CSP loss, and the experimental settings are kept exactly the same. The parameters $\varepsilon$ and $\eta$ in Eq. \eqref{eq_spstd} are set to be learnable with an initial value of 1 and 4. As shown in Table \ref{tab_model} and Fig. \ref{fig_model}, compared to the loss-based approach, TCSP-SAM2 achieves superior performance in almost all evaluation metrics.

\begin{table*}[htbp]
\centering
\caption{Quantitative comparison of SAM2 with CSP loss and TCSP-SAM2.}
\label{tab_model}
\begin{tabular}{lc|ccc|cc|cccc}
\toprule
Dataset & model & Recall $\% \uparrow$ & Dice $\% \uparrow$ & IoU $\% \uparrow$ & HD95 $\downarrow$ & ASSD $\downarrow$ & clDice $\% \uparrow$ & $\beta_0 \downarrow $ & $\beta_1 \downarrow $ & $\chi_{error} \downarrow $ \\
\midrule
\multirow{2}{*}{DCA} & CSP Loss & 83.1488 & 81.5520 & 68.9595 & 9.5617 & 1.9607 & 88.3970 & 5.0000 & 1.0000 & 4.7059 \\
& TCSP-SAM2 & \textbf{84.8496} & \textbf{81.6098} & \textbf{69.0572} & \textbf{8.5621} & \textbf{1.8814} & \textbf{88.8317} & \textbf{0.7353} & \textbf{4.5294} & \textbf{4.3235} \\
\midrule
\multirow{2}{*}{DRIVE} & CSP Loss & 82.6406 & 82.3224 & 69.9762 & 4.5711 & 1.0748 & 82.0546 & 142.3000 & 32.7000 & 175.0000 \\
& TCSP-SAM2 & \textbf{83.6327} & \textbf{82.3911} & \textbf{70.0761} & \textbf{4.4165} & \textbf{1.0619} & \textbf{82.6745} & \textbf{123.5500} & \textbf{31.6500} & \textbf{155.2000} \\
\midrule
\multirow{2}{*}{MASS} & CSP Loss & \textbf{76.7728} & 77.0591 & 63.3338 & 45.9222 & 7.6505 & \textbf{87.5355} & 7.8634 & \textbf{4.0820} & 10.5464 \\
& TCSP-SAM2 & 73.4684 & \textbf{77.5048} & \textbf{64.1304} & \textbf{43.9437} & \textbf{7.1608} & 87.3778 & \textbf{6.4863} & 4.9563 & \textbf{10.3716} \\
\midrule
\multirow{2}{*}{UBW} & CSP Loss & 85.1967 &  84.5309 & 74.2561 & 3.5191 & 1.0265 & 93.6348 & 0.8698 & 0.4010 & 1.0417 \\
& TCSP-SAM2 & \textbf{85.4664} & \textbf{84.7987} & \textbf{74.6871} & \textbf{3.4978} & \textbf{1.0089} & \textbf{93.7596} & \textbf{0.4583} & \textbf{0.2656} & \textbf{0.7031}  \\
\bottomrule
\end{tabular}
\end{table*}

\section{Conclusion and Discussion}\label{sec_conclusion}

In this paper, we address the fundamental challenge of integrating simple-point-based topology-preserving constraints into deep learning frameworks for image segmentation and skeleton extraction. Existing simple point detection methods are limited to binary images, rendering them incompatible with gradient-based optimization or prone to introducing biased gradients, while morphological or purely data-driven approaches often fail to guaranty topological consistency. To overcome these limitations, we propose a method that directly computes simple points on continuous-valued images, enabling differentiable topological inference. Building on this theory, we develop an efficient skeleton extraction algorithm, termed CSPS, which preserves topological properties in both binary and continuous-valued domains. Furthermore, we design a topology-preserving variational model, named TCSP, which enforces topological constraints by retaining the connected components corresponding to topologically non-removable (i.e., non-simple) points. The TCSP model can be seamlessly integrated into any segmentation network with softmax or sigmoid output, as demonstrated by the TCSP-SAM2 architecture. Although our discussion focuses on binary topology preservation, the same principle can be extended to multi-class scenarios. Experimental results demonstrate the effectiveness of the CSP loss in improving the performance of learning-based segmentation networks, as well as the ability of the TCSP model to preserve global topology and geometric structure.

Despite the promising results achieved, several directions remain open for future exploration. The current framework is primarily designed for 2D images. Extending the continuous-tone simple point theory and the corresponding skeleton extraction algorithm to 3D volumetric data presents certain challenges, as the connectivity of 3D neighborhoods cannot be easily determined using crossing numbers in the same way as in the 2D case. Furthermore, beyond image segmentation, the proposed continuous-tone simple point method may also be applied to other image analysis tasks such as image inpainting and image generation, where preserving topological integrity is equally critical.


\bibliographystyle{IEEEtran}
\bibliography{ref.bib}

@InProceedings{deeplabv3+,
author="Chen, Liang-Chieh
and Zhu, Yukun
and Papandreou, George
and Schroff, Florian
and Adam, Hartwig",
title="Encoder-{D}ecoder with Atrous Separable Convolution for Semantic Image Segmentation",
booktitle="Proc. Eur. Conf. Comput. Vis. (ECCV)",
year="2018",
pages="833--851",
month="Oct."
}

@InProceedings{unet,
author="Ronneberger, Olaf
and Fischer, Philipp
and Brox, Thomas",
title="U-{N}et: Convolutional Networks for Biomedical Image Segmentation",
booktitle="Proc. Int. Conf. Med. Image Comput. Comput.-Assist. Interv. (MICCAI)",
year="2015",
pages="234--241",
month="Nov."
}

@ARTICLE{unet++,
  author={Zhou, Zongwei and Siddiquee, Md Mahfuzur Rahman and Tajbakhsh, Nima and Liang, Jianming},
  journal={IEEE Trans. Med. Imaging}, 
  title={U{N}et++: Redesigning Skip Connections to Exploit Multiscale Features in Image Segmentation}, 
  year={2020},
  month={Jun.},
  volume={39},
  number={6},
  pages={1856-1867},
  doi={10.1109/TMI.2019.2959609}}

@inproceedings{segformer,
 author = {Xie, Enze and Wang, Wenhai and Yu, Zhiding and Anandkumar, Anima and Alvarez, Jose M. and Luo, Ping},
 booktitle = {Proc. Adv. Neural Inf. Process. Syst. (NeurIPS)},
 pages = {12077--12090},
 title = {Seg{F}ormer: Simple and Efficient Design for Semantic Segmentation with Transformers},
 volume = {34},
 year = {2021},
 month={Dec.}
}

@INPROCEEDINGS{sam,
  author={Kirillov, Alexander and others},
  booktitle={Proc. IEEE/CVF Int. Conf. Comput. Vis. (ICCV)}, 
  title={Segment Anything}, 
  year={2023},
  month={Oct.},
  volume={},
  number={},
  pages={3992-4003},
  doi={10.1109/ICCV51070.2023.00371}}

@inproceedings{sam2,
title={{SAM} 2: Segment Anything in Images and Videos},
author={Nikhila Ravi and others},
booktitle={Proc. Int. Conf. Learn. Represent. (ICLR)},
year={2025},
month={Apr.},
pages = {41175-41218}
}

@INPROCEEDINGS{snakeconv,
  author={Qi, Yaolei and He, Yuting and Qi, Xiaoming and Zhang, Yuan and Yang, Guanyu},
  booktitle={Proc. IEEE/CVF Int. Conf. Comput. Vis. (ICCV)}, 
  title={Dynamic Snake Convolution based on Topological Geometric Constraints for Tubular Structure Segmentation}, 
  year={2023},
  volume={},
  number={},
  month={Oct.},
  pages={6047-6056},
  doi={10.1109/ICCV51070.2023.00558}}

@InProceedings{fractal,
author="Huang, Jiaxing
and Zhou, Yanfeng
and Luo, Yaoru
and Liu, Guole
and Guo, Heng
and Yang, Ge",
title="Representing Topological Self-similarity Using Fractal Feature Maps for Accurate Segmentation of Tubular Structures",
booktitle="Proc. Eur. Conf. Comput. Vis. (ECCV)",
year="2024",
month="Oct.",
pages="143--160",
doi="10.1007/978-3-031-73404-5_9"
}

@ARTICLE{euler,
  author={Li, Liu and Ma, Qiang and Ouyang, Cheng and Paetzold, Johannes C. and Rueckert, Daniel and Kainz, Bernhard},
  journal={IEEE Trans. Med. Imaging}, 
  title={Topology Optimization in Medical Image Segmentation With Fast {$\chi$} Euler Characteristic}, 
  year={2025},
  month={Jul.},
  volume={44},
  number={12},
  pages={5221-5232},
  doi={10.1109/TMI.2025.3589495}}

@inproceedings{dmtloss,
 author = {Gupta, Saumya and Zhang, Yikai and Hu, Xiaoling and Prasanna, Prateek and Chen, Chao},
 booktitle = {Proc. Adv. Neural Inf. Process. Syst. (NeurIPS)},
 pages = {8186--8207},
 title = {Topology-Aware Uncertainty for Image Segmentation},
 volume = {36},
 year = {2023},
 month = {Nov.}
}

@inproceedings{dmttopologyaware,
title={Topology-Aware Segmentation Using Discrete {M}orse Theory},
author={Xiaoling Hu and Yusu Wang and Li Fuxin and Dimitris Samaras and Chao Chen},
booktitle={Proc. Int. Conf. Learn. Represent. (ICLR)},
year={2021},
month={May},
pages={2101-2119}
}

@INPROCEEDINGS{vggtopo,
  author={Mosinska, Agata and Marquez-Neila, Pablo and Kozinski, Mateusz and Fua, Pascal},
  booktitle={Proc. IEEE/CVF Conf. Comput. Vis. Pattern Recognit. (CVPR)}, 
  title={Beyond the Pixel-Wise Loss for Topology-Aware Delineation}, 
  year={2018},
  month={Jun.},
  volume={},
  number={},
  pages={3136-3145},
  doi={10.1109/CVPR.2018.00331}}

@ARTICLE{structure-agnostic,
  author={Song, Dongning and Huang, Weijian and Liu, Jiarun and Islam, Md Jahidul and Yang, Hao and Wang, Shuqiang and Zheng, Hairong and Wang, Shanshan},
  journal={IEEE Trans. Image Process.}, 
  title={Optimized Vessel Segmentation: A Structure-Agnostic Approach With Small Vessel Enhancement and Morphological Correction}, 
  year={2025},
  month={Oct.},
  volume={34},
  number={},
  pages={7168-7179},
  doi={10.1109/TIP.2025.3607583}}

@INPROCEEDINGS{cldice,
  author={Shit, Suprosanna and others},
  booktitle={Proc. IEEE/CVF Conf. Comput. Vis. Pattern Recognit. (CVPR)}, 
  title={{clDice} - a Novel Topology-Preserving Loss Function for Tubular Structure Segmentation}, 
  year={2021},
  month={Jun.},
  volume={},
  number={},
  pages={16555-16564},
  doi={10.1109/CVPR46437.2021.01629}}

@ARTICLE{learnmorph,
  author={Xie, Jun and Li, Wenxiao and Wang, Faqiang and Zhang, Liqiang and Hou, Zhengyang and Liu, Jun},
  journal={IEEE Trans. Geosci. Remote Sens.}, 
  title={Slender Object Scene Segmentation in Remote Sensing Image Based on Learnable Morphological Skeleton With Segment Anything Model}, 
  year={2025},
  month={Jun.},
  volume={63},
  number={},
  pages={1-13},
  doi={10.1109/TGRS.2025.3581458}}

@InProceedings{clce,
author="Acebes, Cesar
and Moustafa, Abdel Hakim
and Camara, Oscar
and Galdran, Adrian",
title="The Centerline-Cross Entropy Loss for Vessel-Like Structure Segmentation: Better Topology Consistency Without Sacrificing Accuracy",
booktitle="Proc. Int. Conf. Med. Image Comput. Comput.-Assist. Interv. (MICCAI)",
year="2024",
month="Oct.",
pages="710--720",
}

@article{biconnet,
title = {{BiconNet}: An edge-preserved connectivity-based approach for salient object detection},
journal = {Pattern Recogn.},
volume = {121},
pages = {108231},
year = {2022},
month = {Jan.},
issn = {0031-3203},
doi = {10.1016/j.patcog.2021.108231},
author = {Ziyun Yang and Somayyeh Soltanian-Zadeh and Sina Farsiu},
}

@ARTICLE{roadnet,
  author={Liu, Yahui and Yao, Jian and Lu, Xiaohu and Xia, Menghan and Wang, Xingbo and Liu, Yuan},
  journal={IEEE Trans. Geosci. Remote Sens.}, 
  title={{RoadNet}: Learning to Comprehensively Analyze Road Networks in Complex Urban Scenes From High-Resolution Remotely Sensed Images}, 
  year={2019},
  month={Oct.},
  volume={57},
  number={4},
  pages={2043-2056},
  doi={10.1109/TGRS.2018.2870871}}

@INPROCEEDINGS{conn,
  author={Yang, Ziyun and Farsiu, Sina},
  booktitle={Proc. IEEE/CVF Conf. Comput. Vis. Pattern Recognit. (CVPR)}, 
  title={Directional Connectivity-based Segmentation of Medical Images}, 
  year={2023},
  month={Jun,},
  volume={},
  number={},
  pages={11525-11535},
  doi={10.1109/CVPR52729.2023.01109}}

@ARTICLE{affinity,
  author={Oner, Doruk and Koziński, Mateusz and Citraro, Leonardo and Dadap, Nathan C. and Konings, Alexandra G. and Fua, Pascal},
  journal={IEEE Trans. Pattern Anal. Mach. Intell.}, 
  title={Promoting Connectivity of Network-Like Structures by Enforcing Region Separation}, 
  year={2022},
  month={Apr.},
  volume={44},
  number={9},
  pages={5401-5413},
  doi={10.1109/TPAMI.2021.3074366}}

@article{beltrami,
  author    = {Chan, Hei-Long and Yan, Shi and Lui, Lok Ming and Tai, Xue-Cheng},
  title     = {Topology-Preserving Image Segmentation by {B}eltrami Representation of Shapes},
  journal   = {J. Math. Imaging Vis.},
  year      = {2018},
  volume    = {60},
  number    = {3},
  pages     = {401--421},
  month     = {Mar.},
  issn      = {1573-7683},
  doi       = {10.1007/s10851-017-0767-8}}

@article{ZHANG2021218,
title = {Topology- and convexity-preserving image segmentation based on image registration},
journal = {Appl. Math. Model.},
volume = {100},
pages = {218-239},
year = {2021},
month = {Dec.},
doi = {10.1016/j.apm.2021.08.017},
author = {Daoping Zhang and Xue-cheng Tai and Lok Ming Lui}
}

@article{hyperelastic,
  author    = {Zhang, Daoping and Lui, Lok Ming},
  title     = {Topology-Preserving {3D} Image Segmentation Based on Hyperelastic Regularization},
  journal   = {J. Sci. Comput.},
  year      = {2021},
  month     = {Apr.},
  volume    = {87},
  number    = {3},
  pages     = {74},
  issn      = {1573-7691},
  doi       = {10.1007/s10915-021-01433-y}}

@article{quasiconformal,
title = {A learning-based framework for topology-preserving segmentation using quasiconformal mappings},
journal = {Neurocomputing},
volume = {600},
pages = {128124},
year = {2024},
month = {Oct.},
issn = {0925-2312},
doi = {10.1016/j.neucom.2024.128124},
author = {Han Zhang and Lok Ming Lui},
}

@article{deformation,
title = {Anatomically plausible segmentations: Explicitly preserving topology through prior deformations},
journal = {Med. Image Anal.},
volume = {97},
pages = {103222},
year = {2024},
month = {Oct.},
issn = {1361-8415},
doi = {10.1016/j.media.2024.103222},
author = {Madeleine K. Wyburd and Nicola K. Dinsdale and Mark Jenkinson and Ana I.L. Namburete},
}

@article{diffeomorphic,
title = {A fast diffeomorphic image registration algorithm},
journal = {NeuroImage},
volume = {38},
number = {1},
pages = {95-113},
year = {2007},
month = {Oct.},
issn = {1053-8119},
doi = {10.1016/j.neuroimage.2007.07.007},
author = {John Ashburner},
}

@ARTICLE{cloughtopoloss,
  author={Clough, James R. and Byrne, Nicholas and Oksuz, Ilkay and Zimmer, Veronika A. and Schnabel, Julia A. and King, Andrew P.},
  journal={IEEE Trans. Pattern Anal. Mach. Intell.}, 
  title={A Topological Loss Function for Deep-Learning Based Image Segmentation Using Persistent Homology}, 
  year={2022},
  month={Sep.},
  volume={44},
  number={12},
  pages={8766-8778},
  doi={10.1109/TPAMI.2020.3013679}}

@inproceedings{hutopoloss,
 author = {Hu, Xiaoling and Li, Fuxin and Samaras, Dimitris and Chen, Chao},
 booktitle = {Proc. Adv. Neural Inf. Process. Syst. (NeurIPS)},
 pages = {5657-5668},
 title = {Topology-Preserving Deep Image Segmentation},
 volume = {32},
 year = {2019},
 month = {Nov.}
}

@InProceedings{postph,
author="Clough, James R.
and Oksuz, Ilkay
and Byrne, Nicholas
and Schnabel, Julia A.
and King, Andrew P.",
title="Explicit Topological Priors for Deep-Learning Based Image Segmentation Using Persistent Homology",
booktitle="Proc. Inf. Process. Med. Imaging (IPMI)",
year="2019",
pages="16--28",
month="May",
isbn="978-3-030-20351-1",
doi="10.1007/978-3-030-20351-1_2"
}

@misc{widthtopo,
      title={Topology-Guaranteed Image Segmentation: Enforcing Connectivity, Genus, and Width Constraints}, 
      author={Wenxiao Li and Xue-Cheng Tai and Jun Liu},
      year={2026},
      eprint={2601.11409},
      archivePrefix={arXiv},
      primaryClass={cs.CV},
      url={https://arxiv.org/abs/2601.11409}, 
}

@INPROCEEDINGS{topocellgen,
  author={Xu, Meilong and others},
  booktitle={Proc. IEEE/CVF Conf. Comput. Vis. Pattern Recognit. (CVPR)}, 
  title={Topo{C}ell{G}en: Generating Histopathology Cell Topology with a Diffusion Model}, 
  year={2025},
  month={Jun.},
  volume={},
  number={},
  pages={20979-20989},
  doi={10.1109/CVPR52734.2025.01954}}

@ARTICLE{topolevelset,
  author={Xiao Han and Chenyang Xu and Prince, J.L.},
  journal={IEEE Trans. Pattern Anal. Mach. Intell.}, 
  title={A topology preserving level set method for geometric deformable models}, 
  year={2003},
  month={Jun.},
  volume={25},
  number={6},
  pages={755-768},
  doi={10.1109/TPAMI.2003.1201824}}

@article{toposegnet,
title = {TopoSegNet: Scalable topology preservation in image segmentation via critical points},
journal = {Comput. Vis. Image Underst.},
volume = {262},
pages = {104564},
year = {2025},
month = {Dec.},
issn = {1077-3142},
doi = {https://doi.org/10.1016/j.cviu.2025.104564},
author = {Mohsen Ahmadkhani and Eric Shook},
}

@article{topostd,
author = {Deng, Lingyun and Liu, Litong and Wang, Dong and Wang, Xiao-Ping},
title = {Connected-Component Preserving Image Segmentation Using the Iterative Convolution-Thresholding Method},
journal = {SIAM J. Imaging Sci.},
volume = {18},
number = {3},
pages = {1904-1928},
year = {2025},
month={Jul.},
doi = {10.1137/25M1743193},
}

@ARTICLE{deepclose,
  author={Wu, Qian and Chen, Yufei and Liu, Wei and Yue, Xiaodong and Zhuang, Xiahai},
  journal={IEEE Trans. Med. Imaging}, 
  title={Deep Closing: Enhancing Topological Connectivity in Medical Tubular Segmentation}, 
  year={2024},
  month={May},
  volume={43},
  number={11},
  pages={3990-4003},
  doi={10.1109/TMI.2024.3405982}}

@INPROCEEDINGS{spskel,
  author={Menten, Martin J. and others},
  booktitle={Proc. IEEE/CVF Int. Conf. Comput. Vis. (ICCV)}, 
  title={A skeletonization algorithm for gradient-based optimization}, 
  year={2023},
  month={Oct.},
  volume={},
  number={},
  pages={21337-21346},
  doi={10.1109/ICCV51070.2023.01956}}

@misc{ste,
      title={Estimating or Propagating Gradients Through Stochastic Neurons for Conditional Computation}, 
      author={Yoshua Bengio and Nicholas Léonard and Aaron Courville},
      year={2013},
      eprint={1308.3432},
      archivePrefix={arXiv},
      primaryClass={cs.LG},
      url={https://arxiv.org/abs/1308.3432}, 
}

@inproceedings{reparam,
title={Categorical Reparameterization with Gumbel-Softmax},
author={Eric Jang and Shixiang Gu and Ben Poole},
booktitle={Proc. Int. Conf. Learn. Represent. (ICLR)},
year={2017},
month={Apr.},
pages={1920-1931}
}

@article{sp_topo,
title = {Simple points, topological numbers and geodesic neighborhoods in cubic grids},
journal = {Pattern Recogn. Lett.},
volume = {15},
number = {10},
pages = {1003-1011},
year = {1994},
month = {Oct.},
issn = {0167-8655},
doi = {https://doi.org/10.1016/0167-8655(94)90032-9},
author = {Giles Bertrand},
}

@article{MA1996420,
title = {A Fully Parallel 3D Thinning Algorithm and Its Applications},
journal = {Comput. Vis. Image Underst.},
volume = {64},
number = {3},
pages = {420-433},
year = {1996},
month = {Nov.},
issn = {1077-3142},
doi = {https://doi.org/10.1006/cviu.1996.0069},
author = {C.Min Ma and Milan Sonka},
}

@article{ZHAO20071270,
title = {Preprocessing and postprocessing for skeleton-based fingerprint minutiae extraction},
journal = {Pattern Recogn.},
volume = {40},
number = {4},
pages = {1270-1281},
year = {2007},
month = {Apr.},
issn = {0031-3203},
doi = {https://doi.org/10.1016/j.patcog.2006.09.008},
author = {Feng Zhao and Xiaoou Tang},
}

@article{MORSE1994327,
title = {MuItiscale medial analysis of medical images},
journal = {Image Vis. Comput.},
volume = {12},
number = {6},
pages = {327-338},
year = {1994},
month = {Aug.},
issn = {0262-8856},
doi = {https://doi.org/10.1016/0262-8856(94)90057-4},
author = {Bryan S Morse and Stephen M Pizer and Alan Liu},
}

@article{Thibault2000,
  author    = {Thibault, David and Gold, Christopher M.},
  title     = {Terrain Reconstruction from Contours by Skeleton Construction},
  journal   = {GeoInformatica},
  year      = {2000},
  volume    = {4},
  number    = {4},
  pages     = {349--373},
  doi       = {10.1023/A:1026509828354},
  issn      = {1573-7624},
  month     = {Dec.}}

@article{Bertrand2014,
  author    = {Bertrand, Gilles and Couprie, Michel},
  title     = {Powerful Parallel and Symmetric 3D Thinning Schemes Based on Critical Kernels},
  journal   = {J. Math. Imaging Vis.},
  year      = {2014},
  volume    = {48},
  number    = {1},
  pages     = {134--148},
  month     = {Jan.},
  doi       = {10.1007/s10851-012-0402-7},
  issn      = {1573-7683},
}

@article{zhangskel,
author = {Zhang, T. Y. and Suen, C. Y.},
title = {A fast parallel algorithm for thinning digital patterns},
year = {1984},
month = {Mar.},
volume = {27},
number = {3},
issn = {0001-0782},
doi = {10.1145/357994.358023},
journal = {Commun. ACM},
pages = {236–239}}

@ARTICLE{795212,
  author={Zhou, Y. and Toga, A.W.},
  journal={IEEE Trans. Vis. Comput. Graph.}, 
  title={Efficient skeletonization of volumetric objects}, 
  year={1999},
  month={Sep.},
  volume={5},
  number={3},
  pages={196-209},
  doi={10.1109/2945.795212}}

@article{SAHA19971939,
title = {A new shape preserving parallel thinning algorithm for {3D} digital images},
journal = {Pattern Recogn.},
volume = {30},
number = {12},
pages = {1939-1955},
year = {1997},
month = {Dec.},
issn = {0031-3203},
doi = {https://doi.org/10.1016/S0031-3203(97)00016-2},
author = {P.K Saha and B.B Chaudhuri and D {Dutta Majumder}},
}

@ARTICLE{5396343,
  author={Lohou, Christophe and Dehos, Julien},
  journal={IEEE Trans. Pattern Anal. Mach. Intell.}, 
  title={Automatic Correction of {M}a and {S}onka's Thinning Algorithm Using P-Simple Points}, 
  year={2010},
  volume={32},
  number={6},
  month={Jan.},
  pages={1148-1152},
  doi={10.1109/TPAMI.2010.27}}

@ARTICLE{1114851,
  author={Cherng-Min Ma and Shu-Yen Wan and Jiann-Der Lee},
  journal={IEEE Trans. Pattern Anal. Mach. Intell.}, 
  title={Three-dimensional topology preserving reduction on the 4-subfields}, 
  year={2002},
  volume={24},
  number={12},
  pages={1594-1605},
  month={Dec.},
  doi={10.1109/TPAMI.2002.1114851}}

@article{skel_survey,
title = {A survey on skeletonization algorithms and their applications},
journal = {Pattern Recogn. Lett.},
volume = {76},
pages = {3-12},
year = {2016},
month = {Jun.},
issn = {0167-8655},
doi = {https://doi.org/10.1016/j.patrec.2015.04.006},
author = {Punam K. Saha and Gunilla Borgefors and Gabriella {Sanniti di Baja}},
}

@ARTICLE{8000414,
  author={Shen, Wei and Zhao, Kai and Jiang, Yuan and Wang, Yan and Bai, Xiang and Yuille, Alan},
  journal={IEEE Trans. Image Process.}, 
  title={{DeepSkeleton}: Learning Multi-Task Scale-Associated Deep Side Outputs for Object Skeleton Extraction in Natural Images}, 
  year={2017},
  month={Aug.},
  volume={26},
  number={11},
  pages={5298-5311},
  doi={10.1109/TIP.2017.2735182}}

@ARTICLE{9178497,
  author={Liu, Jiang-Jiang and Hou, Qibin and Cheng, Ming-Ming},
  journal={IEEE Trans. Image Process.}, 
  title={Dynamic Feature Integration for Simultaneous Detection of Salient Object, Edge, and Skeleton}, 
  year={2020},
  month={Aug.},
  volume={29},
  number={},
  pages={8652-8667},
  doi={10.1109/TIP.2020.3017352}}

@INPROCEEDINGS{8099896,
  author={Jerripothula, Koteswar Rao and Cai, Jianfei and Lu, Jiangbo and Yuan, Junsong},
  booktitle={Proc. IEEE Conf. Comput. Vis. Pattern Recognit. (CVPR)}, 
  title={Object Co-skeletonization with Co-segmentation}, 
  year={2017},
  month={Jul.},
  volume={},
  number={},
  pages={3881-3889},
  doi={10.1109/CVPR.2017.413}}

@ARTICLE{4766974,
  author={Lobregt, S. and Verbeek, P. W. and Groen, F. C. A.},
  journal={IEEE Trans. Pattern Anal. Mach. Intell.}, 
  title={Three-Dimensional Skeletonization: Principle and Algorithm}, 
  year={1980},
  month={Jan.},
  volume={PAMI-2},
  number={1},
  pages={75-77},
  doi={10.1109/TPAMI.1980.4766974}}

@article{BERTRAND1996115,
title = {A Boolean characterization of three-dimensional simple points},
journal = {Pattern Recogn. Lett.},
volume = {17},
number = {2},
pages = {115-124},
year = {1996},
month = {Feb.},
issn = {0167-8655},
doi = {https://doi.org/10.1016/0167-8655(95)00100-X},
author = {Gilles Bertrand},
}

@inproceedings{BERTRAND1995,
author = {Gilles Bertrand and Zouina Aktouf},
title = {Three-dimensional thinning algorithm using subfields},
volume = {2356},
booktitle = {Vis. Geom. III},
editor = {Robert A. Melter and Angela Y. Wu},
pages = {113 -- 124},
year = {1995},
month = {Jan.},
doi = {10.1117/12.198601},
}

@article{std,
  author    = {Liu, Jun and Wang, Xiangyue and Tai, Xue-Cheng},
  title     = {Deep Convolutional Neural Networks with Spatial Regularization, Volume and Star-Shape Priors for Image Segmentation},
  journal   = {J. Math. Imaging Vis.},
  year      = {2022},
  volume    = {64},
  number    = {6},
  pages     = {625--645},
  month     = {jul.},
  issn      = {1573-7683},
  doi       = {10.1007/s10851-022-01087-x},
}

@ARTICLE{drive,
  author={Staal, J. and Abramoff, M.D. and Niemeijer, M. and Viergever, M.A. and van Ginneken, B.},
  journal={IEEE Trans. Med. Imaging}, 
  title={Ridge-based vessel segmentation in color images of the retina}, 
  year={2004},
  month={Apr.},
  volume={23},
  number={4},
  pages={501-509},
  doi={10.1109/TMI.2004.825627}}

@Article{dca,
AUTHOR = {Cervantes-Sanchez, Fernando and Cruz-Aceves, Ivan and Hernandez-Aguirre, Arturo and Hernandez-Gonzalez, Martha Alicia and Solorio-Meza, Sergio Eduardo},
TITLE = {Automatic Segmentation of Coronary Arteries in {X}-ray Angiograms using Multiscale Analysis and Artificial Neural Networks},
JOURNAL = {Appl. Sci.},
VOLUME = {9},
YEAR = {2019},
month = {Dec.},
NUMBER = {24},
ARTICLE-NUMBER = {5507},
ISSN = {2076-3417},
DOI = {10.3390/app9245507}
}

@phdthesis{mass,
	author = {Mnih, Volodymyr},
	advisor = {Hinton, Geoffrey},
	title = {Machine learning for aerial image labeling},
	year = {2013},
	isbn = {9780494961841},
	school = {University of Toronto},
	address = {CAN.},
	note = {AAINR96184}
}

@proceedings{ubw,
title = {Proceedings of the Third International Symposium on Image Computing and Digital Medicine},
year = {2019},
month = {Aug.},
isbn = {9781450372626},
organization = {Association for Computing Machinery},
address = {New York, NY, USA},
doi = {10.1145/3364836}
}

@article{hd95,
  author    = {Taha, Abdel Aziz and Hanbury, Allan},
  title     = {Metrics for evaluating 3D medical image segmentation: analysis, selection, and tool},
  journal   = {BMC Med. Imaging},
  year      = {2015},
  volume    = {15},
  number    = {1},
  pages     = {29},
  month     = {Aug.},
  doi       = {10.1186/s12880-015-0068-x},
  issn      = {1471-2342},
}

@ARTICLE{assd,
  author={Heimann, Tobias and others},
  journal={IEEE Trans. Med. Imaging}, 
  title={Comparison and Evaluation of Methods for Liver Segmentation From {CT} Datasets}, 
  year={2009},
  month={Feb.},
  volume={28},
  number={8},
  pages={1251-1265},
  doi={10.1109/TMI.2009.2013851}}

@ARTICLE{gudhi,
AUTHOR={Chazal, Frédéric and Michel, Bertrand },  
TITLE={An Introduction to Topological Data Analysis: Fundamental and Practical Aspects for Data Scientists},
JOURNAL={Front. Artif. Intell.},
VOLUME={4},
YEAR={2021},
month={Sep.},
DOI={10.3389/frai.2021.667963},
ISSN={2624-8212}}

@article{skimage,
 title = {scikit-image: image processing in Python},
 author = {van der Walt, Stéfan and others},
 year = {2014},
 month = {Jun.},
 volume = {2},
 pages = {e453},
 journal = {PeerJ},
 issn = {2167-8359},
 doi = {10.7717/peerj.453}
}

@inproceedings{adamw,
title={Decoupled Weight Decay Regularization},
author={Ilya Loshchilov and Frank Hutter},
booktitle={Proc. Int. Conf. Learn. Represent. (ICLR)},
year={2019},
month={May.},
pages={4061-4078},
}

\end{document}